\documentclass[acmsmall, screen]{acmart}

\newif\ifdraft
\draftfalse

\ifdraft
  \newcommand{\todocolor}[1]{\textcolor{red}{#1}}
\else
  \newcommand{\todocolor}[1]{}
\fi
\DeclareRobustCommand{\hlnew}[1]{#1}
\newcommand{\mahmood}[1]{\todocolor{[[Mahmood: #1]]}}
\newcommand{\lujo}[1]{\todocolor{[[Lujo: #1]]}}

\newcommand\todo[1]{\todocolor{[[#1]]}}
\newcommand\nospace[1]{}

\newcommand{\algref}[1]{\mbox{Alg.~\ref{#1}}}
\newcommand{\secref}[1]{\mbox{Sec.~\ref{#1}}\xspace}
\newcommand{\secsref}[2]{\mbox{Sec.~\ref{#1}--\ref{#2}}\xspace}
\newcommand{\figref}[1]{\mbox{Fig.~\ref{#1}}}
\newcommand{\tabref}[1]{\mbox{Table~\ref{#1}}}

\newcommand{\eqnref}[1]{Eqn.~\ref{#1}\xspace}

\makeatletter
\DeclareRobustCommand{\varname}[1]{\begingroup\newmcodes@\mathit{#1}\endgroup}
\makeatother

\newcommand{\frs}[0]{face-rec\-og\-ni\-tion system}
\newcommand{\dnn}[0]{DNN}
\newcommand{\agn}[0]{AGN}
\newcommand{\agns}[0]{\agn{}s}
\newcommand{\vggS}[0]{VGG10}
\newcommand{\vggL}[0]{VGG143}
\newcommand{\vggXL}[0]{VGG2622}
\newcommand{\openfaceS}[0]{OF10}
\newcommand{\openfaceL}[0]{OF143}
\newcommand{\cwloss}[0]{\ensuremath{\Loss_\textit{cw}}}
\newcommand{\subject}[1]{$S_\textit{#1}$}
\newcommand{\dnnFunc}[0]{$F(\cdot)$}
\newcommand{\dnnloss}[0]{$\Loss_\textit{F}$}
\newcommand{\parheading}[1]{\noindent\textbf{#1}~\hspace{2pt}}

\newcommand{\ccsattack}[0]{$\mathit{CCS16}$}
\newcommand{\ccsclipped}[0]{$\widehat{\mathit{CCS16}}$}
\newcommand{\pgd}[0]{$\mathit{PGD}$}

\newcommand{\gan}[0]{GAN}
\newcommand{\gen}[0]{$G$}
\newcommand{\genloss}[0]{$\Loss_\textit{G}$}
\newcommand{\discrim}[0]{$D$}
\newcommand{\discrimgain}[0]{$\Gain_\textit{D}$}
\newcommand{\latent}[0]{$Z$}

\newcommand{\Loss}[0]{\ensuremath{\mathit{Loss}}}
\newcommand{\Gain}[0]{\ensuremath{\mathit{Gain}}}
\newcommand{\data}[0]{\ensuremath{\mathit{data}}}

\usepackage{array}
\usepackage{xspace}
\usepackage{booktabs}
\usepackage{enumerate}
\usepackage{verbatim}
\usepackage{amsmath}
\usepackage{graphicx}
\usepackage{subcaption}
\usepackage{slashbox}
\usepackage[linesnumbered,ruled,noend]{algorithm2e}
\usepackage{siunitx}
\usepackage{setspace} 
\usepackage{pifont}
\usepackage{soul}
\usepackage{doi}
\soulregister\cite7
\soulregister\ref7
\soulregister\pageref7
\soulregister\url7
\soulregister\secref7
\soulregister\figref7
\soulregister\emph7
\soulregister\dnn7
\soulregister\ccsattack7
\soulregister\ccsclipped7
\soulregister\pgd7

\setcopyright{rightsretained}
\acmJournal{TOPS}
\acmYear{2019}
\acmVolume{1}
\acmNumber{1}
\acmArticle{1}
\acmMonth{1}
\acmPrice{}
\acmDOI{10.1145/3317611}

\received{April 2018}
\received[accepted]{March 2019}

\let\log\relax
\DeclareMathOperator{\log}{lg}

\begin{document}

\title{A General Framework for Adversarial Examples with Objectives}


\author{Mahmood Sharif}
\affiliation{%
 \institution{Carnegie Mellon University}
 \city{Pittsburgh}
 \state{PA}
 \country{USA}}
\email{mahmoods@cmu.edu}
\author{Sruti Bhagavatula}
\affiliation{%
 \institution{Carnegie Mellon University}
 \city{Pittsburgh}
 \state{PA}
 \country{USA}}
\email{srutib@cmu.edu}
\author{Lujo Bauer}
\affiliation{%
 \institution{Carnegie Mellon University}
 \city{Pittsburgh}
 \state{PA}
 \country{USA}}
\email{lbauer@cmu.edu}
\author{Michael K. Reiter}
\affiliation{%
 \institution{University of North Carolina at Chapel Hill}
 \city{Chapel Hill}
 \state{NC}
 \country{USA}}
\email{reiter@cs.unc.edu}

\begin{abstract}
  Images perturbed subtly to be misclassified by neural networks, 
  called \textit{adversarial examples}, have emerged as a
  technically deep challenge and an important concern for several
  application domains.  Most research on adversarial examples takes as
  its only constraint that the perturbed images are similar
  to the originals.  However, real-world application of these ideas
  often requires the examples to satisfy additional objectives, which
  are typically enforced through custom modifications of the
  perturbation process.  In this paper, we propose \textit{adversarial
    generative nets} (\agn{}s), a general methodology to train a
  \textit{generator} neural network to emit adversarial examples
  satisfying desired objectives. We demonstrate the ability of
  \agn{}s to accommodate a wide range of objectives, including
  imprecise ones difficult to model, in two application domains.  In particular,
  we demonstrate \textit{physical} adversarial
  examples---eyeglass frames designed to fool face recognition---with
  better robustness, inconspicuousness, and scalability than previous
  approaches, as well as a new attack to fool a handwritten-digit classifier.
\end{abstract}

%
%


\maketitle

\section{Introduction}


Deep neural networks (\dnn{}s) are popular machine-learning models
that achieve state-of-the-art
results on challenging learning tasks in domains where there is
adequate training data and compute power to train them.  For example,
they have been shown to outperform humans in face verification, i.e.,
deciding whether two face images belong to the same
person~\cite{LFW07Tech,Taigman14Deepface}.  Unfortunately, it has also
been shown that \dnn{}s can be easily fooled by \emph{adversarial
  examples}---mildly perturbed inputs that to a human appear visually
indistinguishable from benign inputs---and that such 
adversarial examples can be systematically
found~\cite{Biggio13Evasion,Szegedy13NNsProps}.

While the early attacks and almost all extensions (see
\secref{sec:related}) attempt to meet very few objectives other than
the adversarial input being similar (by some measure) to the original,
there are many contexts in which it is necessary to model additional
objectives of adversarial inputs.  For example, our prior work
considered a scenario in which adversaries could not manipulate input
images directly, but, rather, could only manipulate the physical artifacts
captured in such images~\cite{Sharif16AdvML}.
Using eyeglasses
for fooling face-recognition systems as a driving example, we showed how
to encode various objectives into the process of generating eyeglass
frames, such as ensuring that the frames were capable of being
physically realized by an off-the-shelf printer. As another example, Evtimov et
al.\ considered generating shapes that, when attached to street signs, would
seem harmless to human observers, but would lead neural networks to
misclassify the signs~\cite{Evtimov17Signs}.

These efforts modeled the various objectives they considered in an ad
hoc fashion.
In contrast, in this paper, we propose a \emph{general
framework} for capturing such objectives in the process of generating
adversarial inputs.  Our framework builds on recent work in generative
adversarial networks (\gan{}s)~\cite{Gdfllw14GenAdvNets} to train an attack \textit{generator},
i.e., a neural network that can generate successful attack instances
that meet certain objectives. Moreover, our framework is not only
general, but, unlike previous attacks, produces a large number of
\textit{diverse} adversarial examples that meet the desired
objectives. This could be leveraged by an attacker to generate
attacks that are unlike previous ones (and hence more likely to
succeed), but also by defenders to generate labeled negative inputs to
augment training of their classifiers.  Due to our framework's basis
in \gan{}s, we refer to it using the anagram \agn{}s, for
\textit{adversarial generative nets}.

To illustrate the utility of \agn{}s, we return to the task of
printing eyeglasses to fool face-recognition
systems~\cite{Sharif16AdvML} and demonstrate how to accommodate a
number of types of objectives within it.  Specifically, we use
\agn{}s to accommodate \textit{robustness} objectives to ensure that
produced eyeglasses fool face-recognition systems in different imaging
conditions (e.g., lighting, angle) and even despite the deployment of
specific defenses; \textit{inconspicuousness} objectives, so that the
eyeglasses will not arouse the suspicion of human onlookers; and
\textit{scalability} objectives requiring that relatively few
adversarial objects are sufficient to fool \dnn{}s in many contexts.
We show that \agns{} can be used to target two \dnn{}-based
face-recognition algorithms that achieve human-level
accuracy---VGG~\cite{Parkhi15DeepFR} and
OpenFace~\cite{Amos16Openface}---and output eyeglasses that enable an
attacker to either evade recognition or to impersonate a
specific target, while meeting these additional
objectives. To demonstrate that \agn{}s can be effective in contexts
other than face recognition, we also train \agn{}s to fool a
classifier designed to recognize handwritten digits and trained on the
MNIST dataset~\cite{MNIST}.

In addition to illustrating the extensibility of \agn{}s to various
types of objectives, these demonstrations highlight two additional
features that, we believe, are significant advances.  First, \agn{}s
are \textit{flexible} in that an \agn{} can train a
generator to produce adversarial instances with only vaguely specified
characteristics. For example, we have no way of capturing
inconspicuousness mathematically; rather, we can specify it only using
labeled instances.  Still, \agn{}s can be trained to produce new and
convincingly inconspicuous adversarial examples.  Second, \agn{}s are
\textit{powerful} in generating adversarial examples that perform
better than those produced in previous efforts using more customized
techniques. For example, though some of the robustness and inconspicuousness
objectives we consider here were also considered in prior work, the adversarial
instances produced by \agn{}s perform better (e.g., $\sim$70\% vs.\ 31\%
average success rate in impersonation) and accommodate
other objectives (e.g., robustness to illumination changes).
\agn{}s enable attacks that previous methods did not.

We next review related work (\secref{sec:related}) and then describe the
\agn{} framework (\secref{sec:attack}), and its instantiation
against face-recognition \dnn{}s (\secref{sec:attack-impl}). Then,
we evaluate the effectiveness of \agn{}s, including with physically realized
attacks and a  user study to examine inconspicuousness (\secref{sec:results}).
Finally, we discuss our work and conclude (\secref{sec:conclusion}).

\section{Related Work}
\label{sec:related}

In this section, we describe prior work on test-time attacks on machine-learning
algorithms, work that focuses on physical-world attacks, and proposed defenses.

\parheading{Fooling Machine Learning Algorithms}
In concurrent research efforts, Szegedy et al.\ and Biggio et
al.\ showed how to systematically find adversarial examples to
fool \dnn{}s~\cite{Biggio13Evasion,Szegedy13NNsProps}. Given an input
$x$ that is classified to $F(x)$ by the \dnn{}, Szegedy et al.'s goal was to
find a perturbation $r$ of minimal norm (i.e., as imperceptible as
possible) such that $x+r$ would be classified to a desired target
class $c_t$.  They showed that, when the \dnn{} function, $F(\cdot)$,
and the norm function are differentiable, finding the perturbation can be
formalized as an optimization to be solved
by the L-BFGS solver~\cite{Nocedal80LBFGS}.
\hlnew{Differently from the minimal perturbations proposed by Szegedy et al.,
  Biggio et al.}\ 
\hlnew{focused on finding perturbations that would significantly increase the
  machine-learning algorithm's confidence in the target
  class~\cite{Biggio13Evasion}. Attacks akin to Biggio et al.'s are more suitable for the
  security domain, where one may be interested in assessing the security of
  algorithms or systems under worst-case attacks~\cite{Biggio18Survey,Google18Rules}.}

More efficient algorithms were later proposed for finding adversarial
examples that were even more imperceptible (using different notions
of imperceptibility) or were misclassified with even higher confidence~\cite{Carlini17Robustness,
  Engstrom17AdvTrans, Gdfllw14ExpAdv, Huang15Attack, Kanbak17Transform,
  Miyato16Attack, Moosavi16DeepFool, Papernot16Limitations, Rozsa16Attack,
  Sabour2016Attack}.  
For example, Papernot et al.'s algorithm aims to minimize the number of pixels
changed~\cite{Papernot16Limitations}. Carlini and Wagner experimented
with different formulations of the optimization's objective function
for finding adversarial examples~\cite{Carlini17Robustness}.  They
found that minimizing a weighted sum of the perturbation's norm and a
particular classification loss-function, \cwloss{}, helps achieve more
imperceptible attacks. They defined \cwloss{} not directly over the
probabilities emitted by \dnnFunc{}, but rather over the
\emph{logits}, $L(\cdot)$. The logits are usually the output of the
one-before-last layer of \dnn{}s, and higher logits for a class imply
higher probability assigned to it by the \dnn{}.  Roughly speaking,
\cwloss{} was defined as follows:
$$\cwloss{} = \max\{L_{c}(x+r) : c \neq c_{t}\}-L_{c_{t}}(x+r)$$
where $L_c(\cdot)$ is the logit for class $c$.
Minimizing \cwloss{} increases the probability of the target class, $c_t$, and
decreases the probability of others.

Perhaps closest to our work is the work of Baluja and Fischer~\cite{Fischer17ATNs}.
They propose to train an auto-encoding neural network that takes an image
as input and outputs a perturbed version of the same image that would
be misclassified. Follow-up research efforts concurrent to ours propose to
train generative neural networks to create adversarially perturbed images
that lead to misclassification~\cite{Poursaeed17GANAttack,Xiao18GANAttack,Zhao18GANAttack}.
These attacks require only that the perturbations have a small norm, and
allow perturbations to cover the entire image. In contrast, as we discuss in 
\secsref{sec:attack}{sec:attack-impl}, 
we propose attacks that must satisfy stricter constraints
  (e.g., cover only a small, specific portion of the image) and multiple
  objectives (e.g., generate eyeglasses that both lead to
  misclassification and look realistic).

Moosavi et al.\ showed how to find universal adversarial
perturbations, which lead not just one image to be misclassified, but a large
set of images~\cite{Moosavi17Universal}. Universal perturbations improve our understanding of \dnn{}s' limitations,
as they show that adversarial examples often lie in fixed directions (in the images' RGB
space) with respect to their corresponding benign inputs.
Differently from that work, we explore 
universal attacks that are both inconspicuous and constrained to a small region.

Some work has explored digital-domain attacks that satisfy certain
objectives~\cite{Chen17GraphAttacks,Grosse17Malware,Srndic14PDFfool,Xu16PDFfool}.
For example, Xu et al.\ showed to automatically create malicious PDFs
that cannot be detected using machine-learning
algorithms~\cite{Xu16PDFfool}. These methods use customized algorithms
to achieve very precise objectives (e.g., create a valid PDF output).
Our work instead
focuses on a general method for meeting objectives that might be
imprecise.

Research suggests that adversarial examples are not a result of
overfitting, as in that case adversarial examples would be unlikely to transfer between models
(i.e., to fool models with different architecture or training data than the 
ones the adversarial examples were crafted for)~\cite{Warde16Adv}. A widely held conjecture attributes adversarial
examples to the inflexibility of classification
models~\cite{Fawzi16Robustness,Gdfllw14ExpAdv,Shamir19Explain,Warde16Adv}. 
This conjecture is supported by the success of attacks that approximate \dnn{}s'
classification boundaries by linear separators~\cite{Gdfllw14ExpAdv,Moosavi16DeepFool}.

\parheading{Physical Attacks on Machine Learning}
Kurakin et al.\ demonstrated that imperceptible adversarial examples
can fool \dnn{}s even if the input to the \dnn{} is an image
of the adversarial example printed on paper~\cite{Kurakin16Realize}.
Differently than us, they created adversarial perturbations that covered the
entire image they aimed to misclassify.
Recently, Evtimov et al.\ showed that specially crafted patterns
printed and affixed to street signs can mislead
\dnn{}s for street-sign recognition~\cite{Evtimov17Signs}. Unlike the
work described in this paper, they specified the printability and inconspicuousness objectives
in an ad-hoc fashion.

Our prior work proposed using eyeglasses to perform physically
realizable dodging and impersonation against state-of-the-art \dnn{}s for facial
recognition~\cite{Sharif16AdvML}. The problem of finding adversarial eyeglass
patterns was formalized as an optimization problem with multiple ad hoc
objectives to increase the likelihood that: \emph{1)} face recognition can be fooled 
even when the attacker's pose changes slightly; \emph{2)} transitions between
neighboring pixels on the eyeglasses are smooth; and \emph{3)}
the eyeglasses' colors can be realized using an off-the-shelf printer.
Unlike that prior work, the approach described in this paper is
a general framework for generating adversarial
inputs, in this context instantiated to generate eyeglass patterns as similar
as possible to real designs, while still fooling the \dnn{}s in a
desired manner. Moreover, unlike prior work, we evaluate the
inconspicuousness of the eyeglasses using a user study. We find that the new algorithm can produce more robust
and inconspicuous attacks (\secref{sec:real} and \secref{sec:userstudy}).
We also show the attacks produced by the method we describe here to be
scalable as well as robust in the face of defenses.

Another line of work attempts to achieve privacy from \frs{}s by completely avoiding
\emph{face detection}~\cite{HarveyCvDazz,Privisor13Yamada}. Essentially, face detection
finds sub-windows in images that contain faces, which are later sent for processing by
\frs{}s. Consequently, by evading detection, one avoids the post processing of her
face image by recognition systems. The proposed techniques are not inconspicuous: they either
use excessive makeup~\cite{HarveyCvDazz} or attempt to blind the camera using
light-emitting eyeglasses~\cite{Privisor13Yamada}.

The susceptibility to attacks of learning systems that operate on non-visual input
has also been studied~\cite{Carlini16HVC,Cisse17Houdini,Eberz17ECG,Zhang17Dolphin}.
For instance, researchers showed that speech-recognition
systems can be misled to interpret sounds unintelligible to humans
as actual commands~\cite{Carlini16HVC}. 

\parheading{Defending Neural Networks}
Proposals to ameliorate \dnn{}'s susceptibility to adversarial
examples follow three main directions. One line of work proposes techniques for
training \dnn{}s that would correctly classify adversarial inputs or
would not be susceptible to small perturbations. Such techniques involve
augmenting training with adversarial examples in the hope that
the \dnn{} will learn to classify them
correctly~\cite{Gdfllw14ExpAdv,Kantchelian16ICML,Kannan18ALP,Kurakin16AdvTrain,
Szegedy13NNsProps}. These techniques were found to increase the norms of
the perturbations needed to achieve misclassification. However, it remains unclear 
whether the increase is sufficient to make adversarial examples noticeable to humans.
A recent adversarial training method significantly enhanced the robustness of 
\dnn{}s by training using examples generated with the Projected Gradient
Sign (\pgd{}) attack, which is conjectured to be the strongest attack using local
derivative information about \dnn{}s~\cite{Madry17AdvTraining}.
Two subsequent defenses achieved relatively high success by approximating the
outputs achievable via certain norm-bounded perturbations and
then ensuring these outputs are classified correctly~\cite{Kolter17Defense,Mirman18Defense}.
Unfortunately, the recent
defenses~\cite{Kannan18ALP,Kolter17Defense,Madry17AdvTraining,Mirman18Defense}
are limited to specific types of perturbations (e.g., ones bounded in
$L_{\infty}$),
similarly to their
predecessors~\cite{Gdfllw14ExpAdv,Kantchelian16ICML,Kurakin16AdvTrain,Szegedy13NNsProps}.

A second line of work proposes techniques to detect adversarial examples
(e.g.,~\cite{Feinman17Detector,Grosse17Detector,Meng17Magnet,Metzen17Detector}). The
main assumption of this line of work is that adversarial examples follow a
different distribution than benign inputs, and hence can be
detected via statistical techniques. For instance, Metzen et al.\ propose to
train a neural network to detect adversarial examples~\cite{Metzen17Detector}.
The detector would take its input from an intermediate layer of a \dnn{} and
decide whether the input is adversarial. It was recently shown that this detector,
as well as others, can be evaded using different attack techniques than the
ones on which these detectors were originally evaluated~\cite{Carlini17Bypass}.

A third line of work suggests to transform the \dnn{}s' inputs to
  sanitize adversarial examples and lead them to be correctly classified, while
  keeping the \dnn{}s' original training procedures
  intact~\cite{Guo18ReformDef,Liao18ReformDef,Meng17Magnet,Samangouei18DefGAN,Srini18ReformDef}.
  The transformations aim to obfuscate the gradients on which attacks often
  rely. In certain cases the defenses even rely on undifferentiable
  transformations (e.g., JPEG compression) to prevent the back propagation of
  gradients 
  (e.g.,~\cite{Guo18ReformDef}). Unfortunately, researchers
  have shown that it is possible to circumvent such defenses, sometimes by vanilla
  attacks~\cite{Athalye18BeatCVPR}, and other times by more advanced means
  (e.g., by approximating the input-transformation functions using smooth and
  differentiable functions)~\cite{Athalye18Attack}.

%




\section{A Novel Attack Against \dnn{}s}
\label{sec:attack}

In this section, we describe a new algorithm to
attack \dnn{}s.  We define our threat
model in \secref{sec:threatmodel}, discuss the challenges posed by
vaguely specified objectives in \secref{sec:objectives},
provide background on Generative
Adversarial Networks in \secref{sec:gans}, and
describe the attack framework in \secref{sec:framework}.

\subsection{Threat Model}
\label{sec:threatmodel}

We assume an adversary who gains access to an already trained
\dnn{} (e.g., one trained for face recognition).
The adversary cannot poison the
parameters of the \dnn{} by injecting mislabeled data or altering
training data. Instead, she can only alter the inputs
to be classified.

The attacker's goal is to trick the \dnn{} into misclassifying
adversarial inputs. We consider two variants of this attack.  In \textit{targeted}
attacks, the adversary attempts to trick the \dnn{} to misclassify
the input as a specific class (e.g., to \textit{impersonate} another
subject enrolled in a face-recognition system). In \textit{untargeted}
attacks, the adversary attempts to trick the \dnn{} to misclassify
the input as an arbitrary class (e.g., to \textit{dodge} recognition by
a face-recognition system). 

The framework we propose supports attacks that seek to satisfy a variety
  of objectives, such as 
  maximizing the \dnn{}'s confidence in the target class in impersonation attacks and
  crafting perturbations that are inconspicuous. Maximizing the confidence in the target class
  is especially important in scenarios where strict criteria may be
  used in an attempt to
  ensure security---for instance, scenarios when the confidence must be above a
  threshold, as is used to to prevent false positives in face-recognition
  systems~\cite{Nissenboim10FR}. While inconspicuousness may not be necessary in
  certain scenarios (e.g., unlocking a mobile device via face
  recognition), attacks that are not inconspicuous could easily be
  ruled out in some safety-critical scenarios (e.g.,
  when human operators monitor face-recognition systems at airports~\cite{Steinbuch17JetBlue}).

We assume
a \emph{white-box}
scenario: The adversary knows the feature space (images in RGB
representation, as is typical in \dnn{}s for image classification) and the
architecture and parameters of the system being
attacked. Studying robustness of DNNs
under such
assumptions is standard in the literature (e.g.,~\cite{Carlini17Robustness,Moosavi16DeepFool}).
Moreover, as shown in \secref{sec:transfer} and in prior work (e.g.,~\cite{Papernot17Blackbox}),
black-box attacks can be built from white-box attacks on local substitute-models.
Gradient approximation techniques
could also be used to generalize our proposed method to black-box
settings (e.g.,~\cite{Fred15MI,Narodytska17Blackbox}).

\subsection{Vaguely Specified Objectives}
\label{sec:objectives}

In practice, certain objectives, such as inconspicuousness, may elude
  precise specification. In early stages of our work, while attempting to
  produce eyeglasses to fool face recognition, we attempted multiple ad-hoc
  approaches to enhance the inconspicuousness of the eyeglasses, with limited
  success. For instance, starting from solid-colored eyeglasses in either of the
  RGB or HSV color spaces, we experimented with algorithms that would gradually adjust the colors until
  evasion was achieved, while fixing one or more of the color channels. 
  We also attempted to use Compositional Pattern-Producing Neural Networks~\cite{Stanley07CPPNs}
  combined with an evolutionary algorithm to produce eyeglasses with symmetric
  or repetitive patterns. These approaches had limited success both at
  capturing inconspicuousness (e.g., real eyeglasses do not necessarily have symmetric
  patterns) and at evasion, or failed completely.

Other approaches to improve inconspicuousness such as ensuring that the transitions between
  neighboring pixels are smooth~\cite{Sharif16AdvML} or limiting the extent to
  which pixels are perturbed (see the \ccsattack{} and \ccsclipped{} attacks in
  \secref{sec:results}) had some success at evasion, but resulted in 
  patterns that our user-study participants deemed as distinguishable from real
  eyeglasses' patterns (see \secref{sec:userstudy}). Therefore, instead of pursuing
  such ad hoc approaches to formalize properties that may be insufficient or
  unnecessary for inconspicuousness, in this work we achieve
  inconspicuousness via a general framework that models inconspicuous
  eyeglasses based on many examples thereof, while simultaneously achieving additional
  objectives, such as evasion.

\subsection{Generative Adversarial Networks}
\label{sec:gans}

Our attacks build on Generative Adversarial Networks
(\gan{}s)~\cite{Gdfllw14GenAdvNets} to create accessories (specifically, eyeglasses)
that closely resemble real ones. \gan{}s provide a
framework to train a neural network, termed the \emph{generator} (\gen{}), 
to generate data that belongs to a distribution (close to the real one) that underlies a target
dataset. 
\gen{} maps samples from a distribution, \latent{},
that we know how to sample from (such as $[-1,1]^d$,
i.e., $d$-dimensional vectors
of reals between $-1$ and $1$) to samples from the target
distribution. 

To train \gen{}, another neural network, called the discriminator (\discrim{}), is used.
\discrim{}'s objective is to discriminate between real and generated samples. Thus,
training can be conceptualized as a game with two players, \discrim{} and
\gen{}, in which \discrim{} is trained to emit 1 on real examples 
and 0 on generated samples, and \gen{} is trained to generate outputs that are (mis)classified
as real by \discrim{}. In practice, training proceeds iteratively and alternates
between updating the parameters of \gen{} and \discrim{} via back-propagation. \gen{}
is trained to minimize the following function:
\begin{equation}
\Loss_{G}(Z,D) = \sum_{z \in Z} \log\Big( 1-D \big( G(z) \big) \Big)
\label{eqn:loss_g}
\end{equation}
\genloss{} is minimized when \gen{} misleads \discrim{} (i.e.,
$D(G(z))$ is 1). \discrim{} is trained to maximize the following function:
\begin{equation}
\Gain_D(G,Z,\data) = \sum_{x \in \data}\log\Big( D\big(x\big) \Big) +  
		\sum_{z \in Z} \log\Big( 1-D \big( G(z) \big) \Big)
\label{eqn:gain_d}
\end{equation}
\discrimgain{} is maximized when \discrim{} emits 1 on real samples
and 0 on all others.

Several \gan{} architectures
and training methods have been proposed, including Wasserstein
\gan{}s~\cite{Arjovsky17WGAN} and Deep Convolutional
\gan{}s~\cite{Radford15DCGAN}, on which we build.

\subsection{Attack Framework}
\label{sec:framework}

Except for a few exceptions~\cite{Fischer17ATNs,Evtimov17Signs,
Poursaeed17GANAttack,Sharif16AdvML,Zhao18GANAttack}, in traditional evasion
attacks against \dnn{}s the attacker directly alters benign inputs to maximize or
minimize a pre-defined function related to the desired misclassification
(see~\secref{sec:related}).
Differently from previous attacks, we propose to train
neural networks to generate outputs that can be used to achieve desired evasions
(among other objectives), instead of iteratively tweaking benign inputs to
become adversarial.

More specifically, we propose to train neural networks to generate images of
artifacts (e.g., eyeglasses) that would lead to misclassification.
We require that the artifacts generated by these neural networks resemble a reference
set of artifacts (e.g., real eyeglass designs), as a means to satisfy an objective
that is hard to specify precisely (e.g., inconspicuousness).
We call the neural networks we propose \emph{Adversarial Generative Nets (\agn{}s)}.
Similarly to \gan{}s, \agn{}s are adversarially trained 
against a discriminator to learn how to generate realistic images. 
Differently from \gan{}s, \agn{}s are also
trained to generate (adversarial) outputs that can mislead given
neural networks (e.g., neural networks designed to recognize faces).

Formally, three neural networks comprise an \agn{}: a generator, \gen{};
a discriminator, \discrim{}; and a pre-trained \dnn{} whose classification function
is denoted by \dnnFunc{}. When given an input $x$ to the \dnn{}, \gen{} is trained to
generate outputs that fool \dnnFunc{} and are inconspicuous by minimizing\footnote{We
slightly abuse notation by writing $x+r$ to denote an
image $x$ that is modified by a perturbation $r$. In practice, we use a mask
and set the values of $x$ within the masked region to the exact
values of $r$.}
\begin{equation}
\label{eqn:gen}
\Loss_{G}(Z,D) - \kappa \cdot \sum_{z\in Z} \Loss_{F}(x + G(z))
\end{equation}
We define \genloss{} in the same manner as in~\eqnref{eqn:loss_g}; minimizing it aims to 
generate real-looking (i.e., inconspicuous) outputs that mislead \discrim{}. \dnnloss{} is a
loss function defined over the \dnn{}'s classification function that is
maximized when training \gen{} 
(as $-$\dnnloss{} is minimized). The definition of \dnnloss{} depends on whether the
attacker aims to achieve an untargeted misclassification or a targeted one. For 
untargeted attacks, we use:
$$\Loss_{F}(x+G(z)) = \sum_{i \neq x} F_{c_i}(x+G(z)) - F_{c_x}(x+G(z))$$
while for targeted attacks we use:
$$\Loss_{F}(x+G(z)) = F_{c_t}(x+G(z)) - \sum_{i \neq t} F_{c_i}(x+G(z))$$
where $F_{c}(\cdot)$ is the \dnn{}'s output for class $c$ (i.e., the estimated probability
of class $c$ in case of a \emph{softmax} activation in the last layer).
By maximizing \dnnloss{}, for untargeted attacks, the probability of the correct class
$c_x$ decreases; for targeted attacks,  the probability
of the target class $c_t$ increases. We chose this
definition of \dnnloss{} because we empirically found that it causes 
\agn{}s to converge faster than \cwloss{} or loss functions
defined via cross entropy, as used in prior work~\cite{Carlini17Robustness,Sharif16AdvML}.
$\kappa$ is a parameter that balances the two objectives of \gen{}; we
discuss it further below.

As part of the training process, \discrim{}'s weights are updated to maximize
\discrimgain{}, defined in \eqnref{eqn:gain_d}, to tell apart realistic and generated samples. 
In contrast to \discrim{} and \gen{},  $F(\cdot)$'s weights are unaltered during training
(as attacks should fool the same \dnn{} at test time).

\begin{algorithm}
  \small
 \caption{\agn{} training}\label{alg:agn}

  \SetKwData{Left}{left}\SetKwData{This}{this}\SetKwData{Up}{up}
  \SetKwFunction{Union}{Union}\SetKwFunction{FindCompress}{FindCompress}
  \SetKwInOut{Input}{Input}\SetKwInOut{Output}{Output}

  \Input{$X$, \gen{}, \discrim{}, \dnnFunc{}, $\mathit{dataset}$, $Z$, $N_e$, $s_b$, 
	$\kappa \in \{0,1\}$}
  \Output{Adversarial \gen{}}
  \BlankLine
  \For{$e\leftarrow 1$ \KwTo $N_e$~}{
      create $\varname{mini-batches}$ of size $s_b$ from $\mathit{dataset}$\;
      \For{$\mathit{batch}$ $\in$ $\varname{mini-batches}$}{
        $z\leftarrow s_b$ samples from $Z$\;
        $\mathit{gen}\leftarrow G(z)$\;
        $\mathit{batch}$$\leftarrow$ $\mathit{concat(gen, batch)}$\;
        \If(\tcp*[h]{update \discrim}){even iteration}{
            update \discrim{} by backpropagating $\frac{\partial \Gain_{D}}{\partial \mathit{batch}}$\;
        }\Else(\tcp*[h]{update \gen}){
            \lIf{\dnnFunc{} fooled}{
               \Return{\gen{}}
            }
            $d_1\leftarrow -\frac{\partial \Gain_{D}}{\partial \mathit{gen}}$\;
            $x\leftarrow s_b$ sample images from $X$\;
            $x\leftarrow x + \mathit{gen}$\;
            Compute forward pass $F(x)$\;
            $d_2\leftarrow \frac{\partial \Loss_{F}}{\partial \mathit{gen}}$\;
            $d_1, d_2 \leftarrow \mathit{normalize}(d_1, d_2)$\;
            $d \leftarrow \kappa \cdot d_1 + (1-\kappa) \cdot d_2$\;
            update \gen{} via backpropagating $d$\;
        }
      }
  }
\end{algorithm}

The algorithm for training \agn{}s is provided in \algref{alg:agn}. The algorithm
takes as input a set of benign examples ($X$), a pre-initialized generator and
discriminator, a neural network to be fooled, a dataset of real examples (which the
generator's output should resemble; in our case this is a dataset of eyeglasses), a
function for sampling from \gen{}'s latent space (\latent{}), the maximum number of
training epochs ($N_e$), the mini-batch\footnote{Mini-batch: A subset of samples from the
dataset used to approximate the gradients and compute updates in an iteration of the
algorithm.} size $s_b$, and $\kappa \in [0,1]$.
The result of the training process is an \emph{adversarial generator} that
creates outputs (e.g., eyeglasses) that fool \dnnFunc{}.
In each training iteration, either \discrim{} or \gen{} is updated using a
subset of the data chosen at random. \discrim{}'s weights are updated via gradient ascent
to increase \discrimgain{}; \gen{}'s weights are updated via gradient descent
to minimize~\eqnref{eqn:gen}. To
balance the generator's two objectives, the gradients from \discrimgain{} and
\dnnloss{} are normalized to the lower Euclidean norm of the
two, and then combined into a weighted average
controlled
by $\kappa$. When
$\kappa$ is closer to zero, more weight is given to fooling \dnnFunc{} and less to making the output of \gen{} realistic. Conversely, setting $\kappa$ closer to
one places more weight on increasing the resemblance between \gen{}'s output and
real examples. Training ends when the maximum number of training epochs is
reached, or when \dnnFunc{} is fooled, i.e., when impersonation or dodging 
is achieved.

\section{\agn{}s that Fool Face Recognition}
\label{sec:attack-impl}

We next describe how we trained \agn{}s to generate
inconspicuous, adversarial eyeglasses that can mislead
state-of-the-art \dnn{}s trained to recognize faces. To do so, we
(1) collect a dataset of real eyeglasses; (2) select the architecture of the
generator and the discriminator, and instantiate their weights;
(3) train \dnn{}s that can evaluate the attacks;
and (4) set the parameters for the attacks.

\subsection{Collecting a Dataset of Eyeglasses}
\label{sec:train-data}

A dataset of real eyeglass-frame designs is necessary to train the
generator to create real-looking attacks. We collected such a dataset
using Google's search
API.\footnote{\url{https://developers.google.com/custom-search/}} To
collect a variety of designs, we searched for ``eyeglasses'' and
synonyms (e.g., ``glasses,'' ``eyewear''), sometimes modified by
an adjective, including colors (e.g.,
``brown,'' ``blue''), trends (e.g., ``geek,'' ``tortoise shell''), and
brands (e.g., ``Ralph Lauren,'' ``Prada''). In total, we made 430
unique API queries and collected 26,520 images.

The images we collected were not of only eyeglasses; e.g., we found
images of cups, vases, and logos of eyeglass brands.
Some images were of eyeglasses worn by models or on complex backgrounds.
Such images would hinder
the training process.
Hence, we trained a classifier to detect and keep only images of
eyeglasses over white backgrounds and not worn by models. Using 250
hand-labeled images, we trained a classifier that identified such
images with 100\% precision
and 65\% recall. After applying it to all the images in the dataset,
8,340 images remained. Manually examining a subset of these images
revealed no false positives.

\begin{figure}
\centering
  \includegraphics[width=1in]{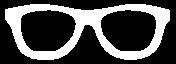}
  \caption{A silhouette of the eyeglasses we use.}
  \label{fig:silhouette}
\end{figure}
\begin{figure}
\centering
  \begin{subfigure}[b]{0.49\textwidth}
    \centering
    \includegraphics[width=0.95in,height=0.35in]{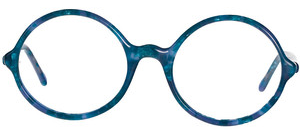}
    \includegraphics[width=0.95in,height=0.35in]{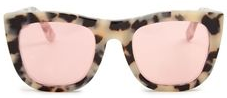}
  \end{subfigure}
  \begin{subfigure}[b]{0.49\textwidth}
    \centering
    \includegraphics[width=0.95in,height=0.35in]{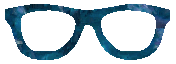}
    \includegraphics[width=0.95in,height=0.35in]{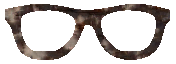}
  \end{subfigure}
  \caption{Examples of raw images of eyeglasses
	that we collected (left) and their synthesis results (right).}
  \label{fig:synth_demo}
\end{figure}

Using images from this dataset, we could
train a generator that can emit eyeglasses of different patterns, shapes, and
orientations. However, variations in shape and orientation made such eyeglasses difficult to
efficiently align to face images while
running~\algref{alg:agn}. Therefore, we preprocessed the images in the
dataset and transferred the patterns from their frames to a fixed shape
(a silhouette of the shape is shown in \figref{fig:silhouette}),
which we could then easily align to face images. 
We then trained the generator to emit images of
eyeglasses with this particular shape, but with different colors and
textures. To transfer the colors and textures of eyeglasses to a fixed
shape, we thresholded the images to detect the areas of the frames.
(Recall that the backgrounds of the images were white.) We then used
Efros and Leung's texture-synthesis technique to synthesize the
texture from the frames onto the fixed
shape~\cite{Efros99Texture}. \figref{fig:synth_demo} shows examples.
Since the texture synthesis process is
nondeterministic, we repeated it twice per image.  At the end of this process, we had 16,680
images for training.

Since physical realizability is a 
requirement for our attacks, it was important that the
generator emitted images of eyeglasses that
are \emph{printable}.  
In particular, the colors of the eyeglasses needed to be
within the range our commodity printer
(see~\secref{sec:real}) could print. Therefore, we  
mapped the colors of the eyeglass frames in the dataset into the color
gamut of our printer. To model the color gamut, we printed an image
containing all $2^{24}$ combinations of RGB triplets, 
captured a picture of that image, and computed the convex hull of
all the RGB triplets in the captured image.
To make an
image of eyeglasses printable, we mapped each RGB triplet in the image
to the closest RGB triplet found within the convex hull.

\begin{figure}[tbh]
	\centering
	\begin{subfigure}[b]{0.99\textwidth}
		\centering
		\includegraphics[height=1in]{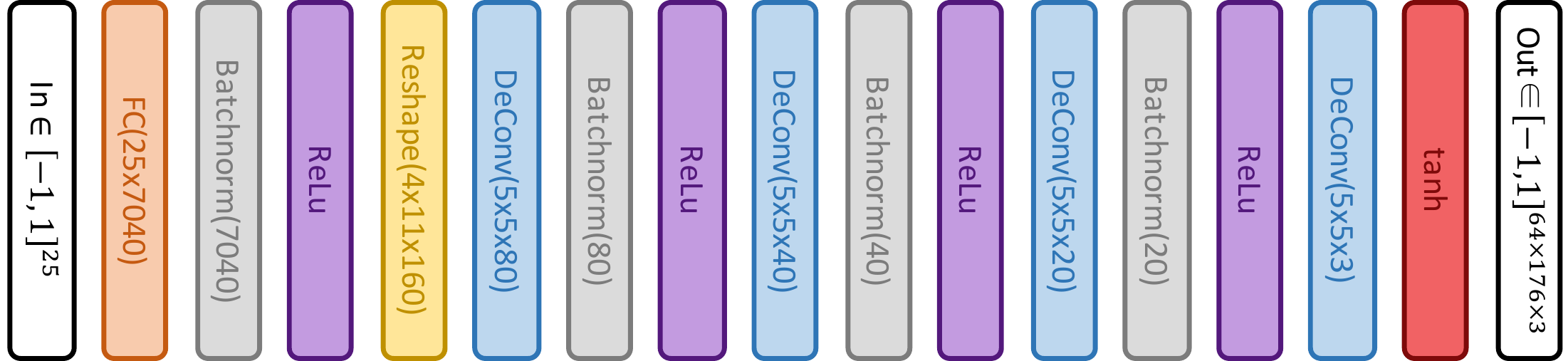}
	\caption{\gen{} (generator).}
	\end{subfigure}
	\begin{subfigure}[b]{0.99\textwidth}
		\centering
		\includegraphics[height=1in]{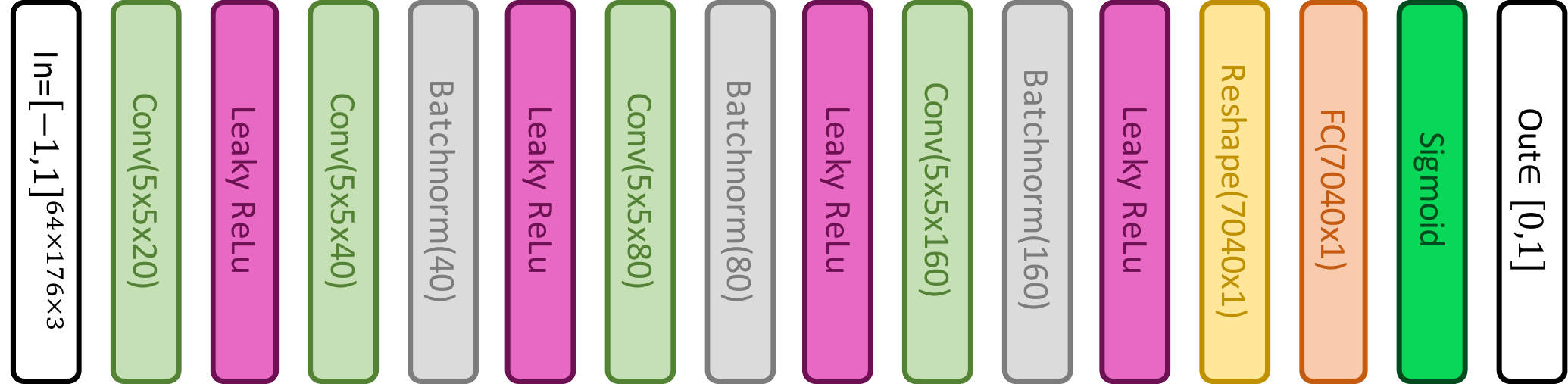}
	\caption{\discrim{} (discriminator).}
	\end{subfigure}
	\begin{subfigure}[b]{0.49\textwidth}
		\centering
		\includegraphics[height=1in]{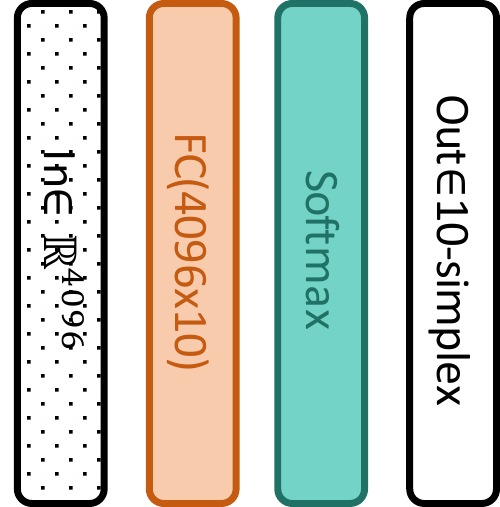}
	\caption{\vggS{}.}
	\end{subfigure}
	\begin{subfigure}[b]{0.49\textwidth}
		\centering
		\includegraphics[height=1in]{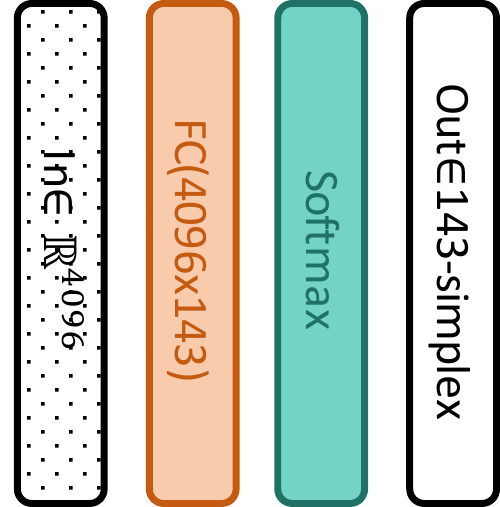}
	\caption{\vggL{}.}
	\end{subfigure}
	\begin{subfigure}[b]{0.49\textwidth}
		\centering
		\includegraphics[height=1in]{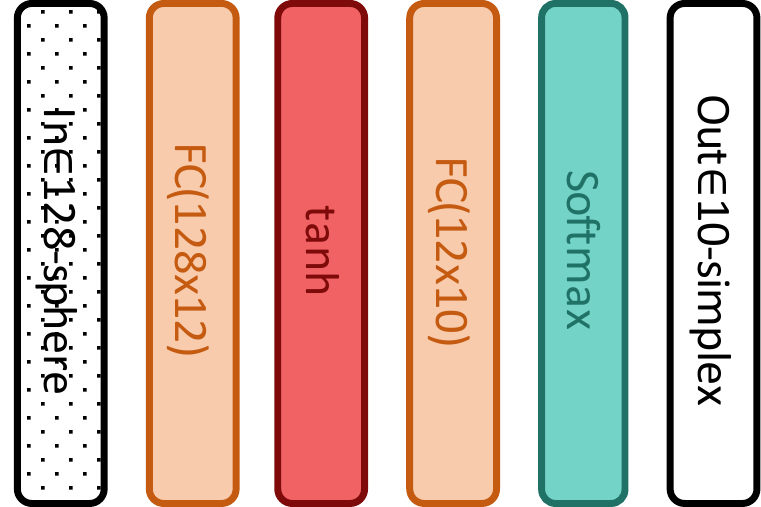}
	\caption{\openfaceS{}.}
	\end{subfigure}
	\begin{subfigure}[b]{0.49\textwidth}
		\centering
		\includegraphics[height=1in]{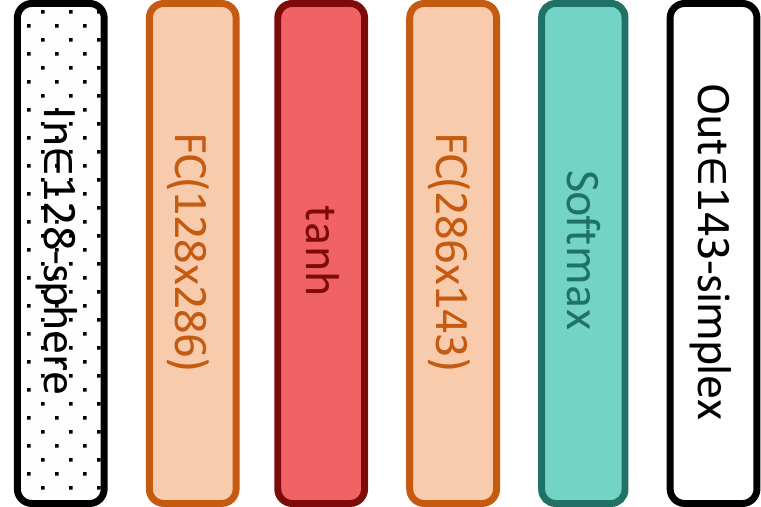}
	\caption{\openfaceL{}.}
	\end{subfigure}
	\begin{subfigure}[b]{0.99\textwidth}
		\centering
		\includegraphics[height=1in]{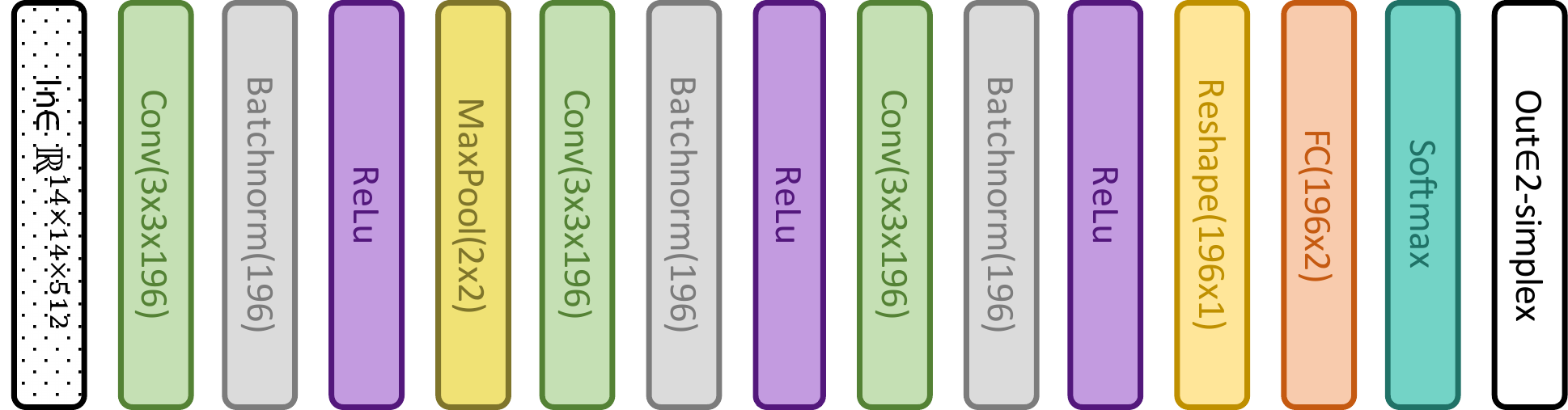}
	\caption{Detector}
	\end{subfigure}
\caption{\label{fig:dnn_archs}
	\hlnew{Architectures of the neural networks used in this work.
	Inputs that are intermediate (i.e., received from feature-extraction
        \dnn{}s) have dotted backgrounds. } 
	\hlnew{
	$\mathit{Deconv}$ refers to transposed convolution,
	and $\mathit{FC}$ to fully connected layer.
	$N$-simplex refers to the set of probability vectors of $N$ dimensions, and the
	128-sphere denotes the set of real 128-dimensional vectors lying on the Euclidean
	unit sphere.
	All convolutions and deconvolutions in \gen{} and \discrim{} have strides and 
	paddings of two.
	The detector's convolutions have strides of two and
	padding of one. The detector's max-pooling layer has a stride of two.}}
\end{figure}


\subsection{Pretraining the Generator and the Discriminator}
\label{sec:pretraining}

When training \gan{}s, it is desirable for the generator to emit
sharp, realistic, diverse images.  Emitting only a small set of images
would indicate the generator's function does not approximate the
underlying distribution well. To achieve these goals, and to enable
efficient training, we chose the Deep Convolutional \gan{}, a minimalistic
architecture with a small number of parameters~\cite{Radford15DCGAN}.
In particular, this architecture is known for its ability to train generators
that can emit sharp, realistic images.

We then explored a variety of options for the
generator's latent space and output dimensionality, as well as the number of
weights in both \gen{} and \discrim{} (via adjusting the depth of
filters). We eventually found that a latent space of $[-1,1]^{25}$
(i.e., 25-dimensional vectors of real numbers between $-1$ and $1$),
and output images of $64 \times 176$ pixels produced the
best-looking, diverse results. The final architectures of \gen{} and
\discrim{} are reported in \figref{fig:dnn_archs}.

\begin{figure}
\centering
  \begin{subfigure}[b]{0.49\textwidth}
    \centering
    \includegraphics[width=0.95in,height=0.35in]{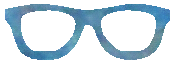}
    \includegraphics[width=0.95in,height=0.35in]{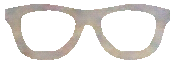}
  \end{subfigure}
  \begin{subfigure}[b]{0.49\textwidth}
    \centering
    \includegraphics[width=0.95in,height=0.35in]{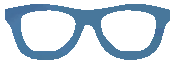}
    \includegraphics[width=0.95in,height=0.35in]{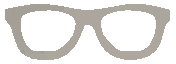}
  \end{subfigure}
  \caption{Examples of eyeglasses emitted by the generator (left)
    and similar eyeglasses from the training set (right).}
  \label{fig:gen_demo}
\end{figure}

To ensure that attacks converged quickly, we initialized \gen{}
and \discrim{} to a state in which the generator can already produce
real-looking images of eyeglasses. To do so, we pretrained \gen{} and
\discrim{} for 200 epochs and stored them to initialize later runs of
\algref{alg:agn}.\footnote{For training, we used the Adam
optimizer~\cite{Kingma14Adam} and set the learning rate to \num{2e-4},
the mini-batch size to 260, $\beta_1$ to 0.5, and $\beta_2$
to 0.999.}
Moreover, we used Salimans et al.'s recommendation
and trained \discrim{} on \emph{soft
  labels}~\cite{Salimans16TrainGANs}. Specifically, we trained
\discrim{} to emit 0 on samples originating from the generator, and
0.9 (instead of 1) on real examples.
\figref{fig:gen_demo} presents a couple of eyeglasses
emitted by the pretrained generator.

\subsection{\dnn{}s for Face Recognition}
\label{sec:dnns}

We evaluated our attacks against four \dnn{}s of two architectures.
Two of the \dnn{}s were built on the \emph{Visual Geometry Group
  (VGG)} neural network~\cite{Parkhi15DeepFR}.
The original VGG \dnn{} 
exhibited
state-of-the-art results on the Labeled Faces in
the Wild (LFW) benchmark, with 98.95\% accuracy for face
verification~\cite{LFW07Tech}. The VGG architecture
contains a large number of weights (the original \dnn{}
contains about $268.52$ million parameters). 
The other two \dnn{}s were built on the
OpenFace neural network, which uses the Google FaceNet
architecture~\cite{Amos16Openface}. OpenFace's main design
consideration
is to provide high accuracy with low training and
prediction times so that the \dnn{} can be deployed on mobile and
IoT devices. Hence, the \dnn{} is relatively compact, with $3.74$
million parameters, but nevertheless
achieves near-human accuracy on the LFW benchmark (92.92\%).

We trained one small and one large face-recognition \dnn{} 
for each architecture. Since we wanted to experiment
with physically realizable dodging and impersonation, we trained the
\dnn{}s to recognize a mix of subjects available to us locally
and celebrities of whom we could acquire images for
training.  The small \dnn{}s were trained to recognize five subjects
from our research group (three females and two males) and five
celebrities from the PubFig dataset~\cite{Kumar2009Attribute}: Aaron
Eckhart, Brad Pitt, Clive Owen, Drew Barrymore, and Milla Jovovich. We
call the small \dnn{} of the VGG and OpenFace architectures \vggS{}
and \openfaceS{}. The large \dnn{}s, termed \vggL{} and
\openfaceL{}, 
were trained to recognize 143 subjects. Three of the subjects were
members of our group, and 140 were celebrities with images in
PubFig's evaluation set. In training, we used about 40 images per
subject. 

\parheading{Training the VGG networks} The original VGG network takes
a $224\times 224$ aligned face image as input and produces a highly
discriminative face descriptor (i.e., vector representation of the
face) of 4096 dimensions. Two descriptors of images of the same person
are designed to be closer to each other in Euclidean space than two
descriptors of different people's images.  We used the descriptors to
train two simple neural networks that map face descriptors to
probabilities over the set of identities. In this manner, the original
VGG network effectively acted as a feature extractor. This is a
standard training approach, termed \emph{transfer learning}, which
is commonly used to train high-performing \dnn{}s using ones that
have already been trained to perform a related task~\cite{Yosinski14TL}.

The architectures of the VGG-derived neural networks 
are provided in~\figref{fig:dnn_archs}. They
consist of fully connected layers (i.e., linear separators)
connected to a \emph{softmax} layer that turns the linear
separators' outputs into probabilities. We trained the networks using
the standard technique of minimizing cross-entropy
loss~\cite{DeepLearningBook16}. After training, we connected the
trained neural networks to the original VGG network to construct end-to-end
\dnn{}s that map face images to identities.

An initial evaluation of \vggS{} and \vggL{} 
showed high
performance. To verify that the \dnn{}s cannot
be easily misled, we tested them against na\"{i}ve attacks by attaching eyeglasses
emitted by the pretrained (non-adversarial) generator to test images. We found that
impersonations of randomly picked targets are unlikely---they occur
with 0.79\% chance for \vggS{} and $<$0.01\% for \vggL{}.
However, we found that dodging would succeed with non-negligible
chance: 7.81\% 
of the time against \vggS{} and 26.87\% against
\vggL{}. We speculated that this was because the training samples for
some subjects never included eyeglasses.
To make the \dnn{}s more robust, we augmented
their training data following adversarial training techniques~\cite{Kurakin16AdvTrain}:
for each image initially used in training we added two
variants with generated eyeglasses attached. We also experimented with using
more variants but found no additional improvement.
Also following Kurakin et al., we included 50\%
raw training images and 50\% augmented images in each mini-batch
during training~\cite{Kurakin16AdvTrain}.

\begin{table}
\begin{center}
\small
  \begin{tabular} {c r r r r r r}
	\toprule

& & \emph{SR na\"{i}ve} & \emph{SR na\"{i}ve}\\
	\emph{Model} & \emph{acc.}  & \emph{dodge} & \emph{impers.} 
		& \emph{thresh.} & \emph{TPR} & \emph{FPR} \\ \midrule
	\vggS & 100\% & 3\% & 0\% & 0.92 & 100\% & 0\% \\
	\vggL & 98\% & 5\% & 0\% & 0.82 & 98\% & 0\% \\
	\openfaceS & 100\% & 14\% & 1\% & 0.55 & 100\% & 0\% \\
	\openfaceL & 86\% & 22\% & $<$1\% & 0.91 & 59\% & 2\% \\
	\bottomrule
  \end{tabular}

	\caption{Performance of the face-recognition \dnn{}s. We report the accuracy,
	the success rate (SR) of na\"{i}ve dodging
	and impersonation (likelihood of na\"{i}ve attackers to
	be misclassified arbitrarily or as \emph{a priori} chosen targets),
	the threshold to balance correct and false classifications, the true-positive
	rate (TPR; how often the correct class is assigned a probability above
	the threshold), and the false-positive rate (FPR; how often a wrong class is
	assigned a probability above the threshold).
	} 
	\label{tab:dnn_performance}
\end{center}
\end{table}

Evaluating \vggS{} and \vggL{} on held-out test sets after training,
we found that they achieved 100\% and 98\% accuracy, respectively. In
addition, the success of na\"ive dodging was at most 4.60\% and
that of impersonation was below 0.01\%.
Finally, to maintain a high level of security, it is
important to minimize the \dnn{}s' false
positives~\cite{Nissenboim10FR}. One way to do so is by setting a
criteria on the \dnn{}s' output to decide when it should be accepted.
We were able to find thresholds for the probabilities emitted by
\vggS{} and \vggL{} such that their accuracies remained
100\% and 98\%, while the false-positive rates of both \dnn{}s were
0\%. The performance of the \dnn{}s 
is reported in \tabref{tab:dnn_performance}.

\parheading{Training the OpenFace networks} The original OpenFace
network takes a $96\times 96$ aligned face image as input and outputs
a face descriptor of 128 dimensions. Similar to the VGG networks, the
descriptors of images of the same person are close in Euclidean space,
while the descriptors of different people's images are far. Unlike
VGG, the OpenFace descriptors lie on a unit sphere.

We again used transfer learning to train the OpenFace networks.
We first attempted to train neural networks that map the OpenFace
descriptors to identities using architectures similar to the ones used
for the VGG \dnn{}s.  We found these neural networks to achieve
competitive accuracies. Similarly to the VGG \dnn{}s, they
were also vulnerable to na\"{i}ve dodging attempts, but 
unlike the VGG \dnn{}s, straightforward data augmentation did not
improve their robustness. We believe this may stem from
limitations of classifying data on a sphere using linear separators.

To improve the robustness of the DNNs, we increased their
depth by prepending a fully connected layer followed by a
hyperbolic-tangent (\emph{tanh}) layer (see \figref{fig:dnn_archs}).
This architecture was chosen as it performed the best out of different
ones we experimented with. We also increased the number of images we
augmented in training to 10 (per image in the training set)
for \openfaceS{} and to 100 for \openfaceL{}. The number of images
augmented was selected such that increasing it did not further improve
robustness against na\"{i}ve attacks.
Similarly to the VGG networks, we trained with about 
 40 images per subject, and
included 50\% raw images and 50\% augmented images in training
mini-batches.

We report the performance of the networks in
\tabref{tab:dnn_performance}. \openfaceS{} achieved 100\% accuracy,
while \openfaceL{} achieved 85.50\% accuracy
(comparable to Amos et al.'s finding~\cite{Amos16Openface}). The OpenFace \dnn{}s
were more vulnerable to na\"{i}ve attacks than the VGG \dnn{}s.
For instance, \openfaceS{} failed against 14.10\% of the na\"{i}ve dodging
attempts and 1.36\% of the na\"{i}ve impersonation attempts.
We believe that the lower accuracy and higher susceptibility of the OpenFace
\dnn{}s compared to the VGG \dnn{}s may stem from the limited capacity of the
OpenFace network induced by the small number of parameters. 

\parheading{Training an attack detector} In addition to the
face-recognition DNNs, we trained a \dnn{} to detect attacks 
that target the VGG networks following the proposal of Metzen et
al.~\cite{Metzen17Detector}.  We chose this detector because it was
found to be one of the most effective detectors against imperceptible adversarial
examples~\cite{Metzen17Detector,Carlini17Bypass}. We focused on
VGG \dnn{}s because no detector architecture was proposed
for detecting attacks against OpenFace-like
architectures.  To mount a successful attack when a detector is
deployed, it is necessary to simultaneously fool the detector and
the face-recognition \dnn{}.

We used the architecture proposed by Metzen et al.\ (see \figref{fig:dnn_archs}). 
For best performance, we attached the detector after the fourth max-pooling
layer of the VGG network. To train the detector, we used 170 subjects from the original
dataset used by Parkhi et al.\ for training the VGG
network~\cite{Parkhi15DeepFR}.
For each subject we used 20 images for
training. For each training image, we created a corresponding
adversarial image that evades recognition. We trained the
detector for 20 epochs using the Adam optimizer
with the training parameters set to standard
values (learning rate $=\num{1e-4}$, $\beta_1=0.99$, $\beta_2=0.999$)~\cite{Kingma14Adam}.
At the end of training, we evaluated the detector on 20
subjects who were not used in training, finding that it had 100\% recall
and 100\% precision.

\subsection{Implementation Details}

We used the Adam optimizer to update the weights of
\discrim{} and \gen{} when running \algref{alg:agn}. As
in pretraining, $\beta_1$ and $\beta_2$ were set to 0.5 and 0.999.
We ran grid search to set $\kappa$ and the learning rate, and found
that a $\kappa=0.25$ and a learning rate of $\num{5e-5}$ gave the best
tradeoff between success in fooling the \dnn{}s, inconspicuousness,
and the algorithm's run time. The number of epochs was limited to
at most one, as we found that the results the algorithm returned when
running longer were not inconspicuous.

The majority of our work was implemented
in MatConvNet, a MATLAB toolbox for convolutional neural
networks~\cite{Vedaldi15Matconvnet}. The OpenFace \dnn{}
was translated from the original implementation in Torch
to MatConvNet. We released the implementation online:
\url{https://github.com/mahmoods01/agns}.


\section{Evaluating \agn{}s}
\label{sec:results}

We extensively evaluated \agn{s} as an attack method.  In
\secref{sec:digital} we show that \agn{}s reliably generate successful
dodging and impersonation attacks in a digital environment, even when
a detector is used to prevent them.  We show in \secref{sec:real} that
these attacks can also be successful in the physical domain.
In \secref{sec:universal}, we demonstrate universal dodging, i.e.,
generating a small number of eyeglasses
that many subjects can use to evade recognition.  We test in
\secref{sec:transfer} how well our attacks transfer between
models. \tabref{tab:exp_obj} summarizes the combinations of domains
the attacks were performed in, the attack types considered, and
what objectives were aimed or tested for in each experiment. 
While more combinations exist, we attempted to experiment with the most interesting
combinations under computational and manpower constraints. (For example,
attacks in the physical domain require significant manual effort, and universal
attacks require testing with more subjects than is feasible to test with in the
physical domain.) In \secref{sec:userstudy}
we demonstrate that \agn{}s can generate eyeglasses that are
inconspicuous to human participants in a user study.
Finally, in \secref{sec:mnist} we show that \agn{}s are applicable
to areas other than face recognition (specifically, by fooling a
digit-recognition \dnn{}).

\newcommand{\cmark}{\ding{51}}

\begin{table}[t]
\begin{center}
\small
  \tabcolsep=0.11cm
  \begin{tabular} { c | c c | c c | c | c c c c c | c | c }
        \toprule
        & \multicolumn{2}{c|}{\textbf{\textit{Domain}}} & \multicolumn{2}{c|}{\textbf{\textit{Type}}} & \multicolumn{8}{c}{\textbf{\textit{Objectives}}} \\
	& & & & 
	&	&  \multicolumn{5}{c|}{\emph{Robustness vs.}}  & \\
        \emph{Sec.} & \emph{Dig.} & \emph{Phys.} & \emph{Untarg.}
	& \emph{Targ.} & \emph{Inconspicuous.} & \emph{Aug.}
	&  \emph{Detect.}& \emph{Print.} & \emph{Pose}
	& \emph{Lum.} & \emph{Universal.} & \emph{Transfer.} \\ \midrule
	\ref{sec:digital} 	& \cmark & & \cmark & & \cmark & \cmark & & \cmark & & & & \\
				 	& \cmark & & \cmark & & \cmark & \cmark & \cmark & \cmark & & & & \\
					& \cmark & & & \cmark & \cmark & \cmark & & \cmark & & & & \\
				 	& \cmark & & & \cmark & \cmark & \cmark & \cmark & \cmark & & & & \\ \midrule
	\ref{sec:real} 		& & \cmark & \cmark & & \cmark & \cmark &  & \cmark & \cmark & \cmark & & \\
					& & \cmark &  & \cmark & \cmark & \cmark &  & \cmark & \cmark & \cmark & & \\ \midrule
	\ref{sec:universal} 	& \cmark & & \cmark & & \cmark & \cmark & & \cmark & & & \cmark & \\ \midrule
	\ref{sec:transfer} 	& \cmark & & \cmark & & \cmark & \cmark & & \cmark & & & & \cmark \\
					& \cmark & & \cmark & & \cmark & \cmark & & \cmark & & & \cmark & \cmark \\
					& & \cmark & \cmark & & \cmark & \cmark & & \cmark & \cmark & \cmark & & \cmark \\
	\bottomrule
	\end{tabular}
  \caption{\label{tab:exp_obj}
	For each experimental section, we mark the combinations of domains
	(digital or physical) in which the attack was tested, the attack types (untargeted
	or targeted) tested, and the objectives of
	the attack, chosen from inconspicuousness, robustness (against training-data
	augmentation, detection, printing noise, pose changes, and luminance
	changes), universality, and transferability. Note that while we did not
	explicitly design the attacks to transfer between architectures, we found that
	they transfer relatively well; see \secref{sec:transfer}.}
\end{center}
\end{table}

\subsection{Attacks in the Digital Domain}
\label{sec:digital}

In contrast to physically realized attacks, an attacker in the digital
domain can exactly control the input she provides to \dnn{}s, since the inputs are not subject to noise added by
physically realizing the attack or capturing the image with a
camera. Therefore, our first step is to verify that the attacker
can successfully fool the \dnn{}s in the digital domain, as failure in
the digital domain implies failure in the physical domain.

\parheading{Experiment setup}
To evaluate the attacks in the digital domain, we selected a set of
subjects for each \dnn{} from the subjects the \dnn{}s were trained
on: 20 subjects selected at random for \vggL{} and \openfaceL{}
and all ten subjects for \vggS{} and \openfaceS{}. 
In impersonation attacks, the targets were chosen at random.
To compute the uncertainty in our estimation of success, we repeated each
attack three times, each time using a different image of the attacker.

\hlnew{As baselines for comparison, we evaluated three additional
  attacks using the same setup. The first attack, denoted
  \ccsattack{},
  is our proposed attack from prior work~\cite{Sharif16AdvML}. The
  \ccsattack{} attack iteratively modifies the colors of the eyeglasses until
  evasion is achieved. We ran the attack up to 300 iterations, starting from solid colors,
  and clipping the colors of the eyeglasses to the range $[0,1]$ after each
  iteration to ensure that they lie in a valid range. The second attack is the
  \pgd{} attack of Madry et al.~\cite{Madry17AdvTraining},
  additionally constrained to
  perturb only the area covered by the eyeglasses. Specifically, we started from
  eyeglasses with solid colors and iteratively perturbed them for up to 100
  iterations, while clipping the perturbations to have max-norm (i.e.,
  $L_\infty$-norm) of at most 0.12. We picked 100 and 0.12 as
  the maximum number of iterations and max-norm threshold, respectively, as these
  parameters led to the most powerful attack in prior
  work~\cite{Madry17AdvTraining}. The \pgd{} attack can be seen as a special
  case of the \ccsattack{} attack, as the two attacks follow approximately the same
  approach except that \pgd{} focuses on a narrower search space by clipping perturbations more aggressively
  to decrease their perceptibility.
  As we show below, the success rate of \pgd{}
  was significantly lower than of \ccsattack{}, suggesting that the clipping \pgd{} performs may
  be too aggressive for this application. Therefore, we evaluated a third attack, denoted
  \ccsclipped{}, in which we clipped perturbations, but did so less aggressively
  than \pgd{}. In \ccsclipped{}, we set the number of iterations to 300, as in
  the \ccsattack{} attack, and clipped perturbations to have max-norm of at most 0.47. We
  selected 0.47 as the max-norm threshold as it is the lowest threshold that led
  to success rates at fooling face recognition that are comparable to
  \ccsattack{}. In other words, this threshold gives the best chance
  of achieving inconspicuousness without substantially sacrificing
  the success rates. To maximize the success rates of the three attacks, we 
  optimized only for the evasion objectives (defined via categorical 
  cross-entropy), and ignored other objectives that are necessary to physically
  realize the attacks (specifically, generating colors that can be printed and 
  ensuring smooth transitions between neighboring pixels~\cite{Sharif16AdvML}).}

To test whether a detector would prevent attacks based on \agn{}s, we selected
all ten subjects for \vggS{} and 20 random subjects for \vggL{}, each
with three images per subject. We then tested whether dodging 
and impersonation can be achieved while simultaneously evading the
detector. To fool the detector along with the face-recognition
\dnn{}s, we slightly modified the objective from \eqnref{eqn:gen} to
optimize the adversarial generator such that the detector's loss is
increased. In particular, the loss function we used was the difference
between the probabilities of the correct class (either ``adversarial''
or ``non-adversarial'' input) and the incorrect class. As the vanilla \pgd{},
\ccsattack{}, and \ccsclipped{} attacks either failed to achieve success
rates comparable to \agn{}s or produced more conspicuous eyeglasses
(\secref{sec:userstudy}), we did not extend them to fool the detector.

We measured the \emph{success rate} of
attacks via two metrics. For dodging, we measured the percentage of
attacks in which the generator emitted
eyeglasses that (1) led the image to be misclassified (i.e., the most probable class was not the attacker), and
(2) kept the 
probability of the correct class below 0.01 (much lower than
the thresholds set for accepting any of the \dnn{}s'
classifications; see \tabref{tab:dnn_performance}). 
For impersonation, we considered an attack successful if the attacker's image
was classified as the target with 
probability exceeding 0.92, the highest
threshold used by any of the \dnn{}s. 
When using the detector, we also required that the detector
deemed the input non-adversarial with probability higher than 0.5.

\begin{table}[t]
\begin{center}
\small
  \tabcolsep=0.11cm
  \begin{tabular} { c | c c c c | c c c c | c c }
	\toprule
	&  \multicolumn{8}{c|}{\emph{Without detector}} 
		&  \multicolumn{2}{c}{\emph{With detector}} \\
        & \multicolumn{4}{c|}{\emph{Dodging}} & \multicolumn{4}{c|}{\emph{Impersonation}}
		& \emph{Dodging} & \emph{Impers.}\\
	\emph{Model} & \emph{\agn{}s} & \pgd{} & \ccsclipped{} & \ccsattack{}
				& \emph{\agn{}s} & \pgd{} & \ccsclipped{} & \ccsattack{}
				& \emph{\agn{}s} & \emph{\agn{}s} \\ \midrule
	\vggS 	& 100$\pm$0\% & 37$\pm$10\% & 100$\pm$0\% & 100$\pm$0\%
		  	& 100$\pm$0\% & 10$\pm$5\% & 100$\pm$0\% & 100$\pm$0\%
			& 100$\pm$0\% & 100$\pm$0\% \\
	\vggL 	&  100$\pm$0\% & 53$\pm$9\% & 100$\pm$0\% & 100$\pm$0\%
			& 88$\pm$5\% & 3$\pm$2\% & 82$\pm$5\% & 98$\pm$2\%
			& 100$\pm$0\% & 90$\pm$4\% \\
	\openfaceS 	& 100$\pm$0\% & 43$\pm$11\% & 100$\pm$0\% & 100$\pm$0\%
				& 100$\pm$0\% & 13$\pm$7\% & 87$\pm$9\% & 100$\pm$0\%
				& - & - \\
	\openfaceL 	& 100$\pm$0\% & 85$\pm$7\% & 100$\pm$0\% & 100$\pm$0\%
				& 90$\pm$4\% & 11$\pm$4\% & 78$\pm$6\% & 88$\pm$4\%
				& - & - \\
	\bottomrule
	\end{tabular}

  \caption{\hlnew{Results of attacks in the digital environment.
	We report the the mean success rate of attacks and the standard error
        when fooling the facial-recognition \dnn{}s. 
	To the right, we report the mean success rates and standard errors of
	dodging and impersonation using \agn{}s when simultaneously fooling
	facial-recognition \dnn{}s and a detector (see~\secref{sec:dnns})}.
	}
	\label{tab:digital}
\end{center}
\end{table}

\begin{figure}
\centering
\includegraphics[width=1in]{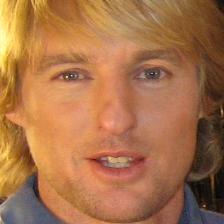}\hspace{1pt}
\includegraphics[width=1in]{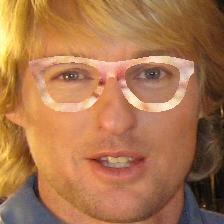}
\caption{An example of digital dodging.
Left: An image of actor Owen Wilson (from the PubFig
dataset~\cite{Kumar2009Attribute}), correctly classified by 
\vggL{} with probability 1.00.
Right: Dodging against \vggL{} using \agn{}'s output (probability
assigned to the correct class $<$0.01).}
\label{fig:digital_dodge}
\end{figure}

\parheading{Experiment results} \tabref{tab:digital} summarizes the results
of the digital-environment experiments. All dodging
attempts using \agn{}s succeeded; \figref{fig:digital_dodge} shows an
example. As with dodging, all impersonation attempts using \agns{} against the
small \dnn{}s (\vggS{} and \openfaceS{}) succeeded.
A few attempts against the larger \dnn{}s failed, suggesting that
inconspicuous impersonation attacks may be more challenging 
when the \dnn{} recognizes many subjects, although attacks succeeded
at least 88\% of the time.

\hlnew{Differently from \agn{}s, the success of \pgd{} was limited---its success
rates at dodging ranged from 37\% to 85\%, and its success
rates at impersonation were below 13\%. This suggests that the
clipping performed by \pgd{} may be too aggressive. The \ccsattack{} and
\ccsclipped{} attacks achieved success rates comparable to \agn{}s at both dodging
and impersonation. Unlike for \agns{}, in this set of
experiments the high success rates of the \ccsattack{} and \ccsclipped{} attacks were
achieved by only attempting to dodge or impersonate (i.e., fool the classifier), while ignoring other
objectives, such as generating colors which can be printed (previously achieved by
minimizing the so-called non-printability
score~\cite{Sharif16AdvML}). For physically realized attacks,
evaluated in~\secref{sec:real}, satisfying these additional
objectives is necessary for the \ccsattack{} and \ccsclipped{}
attacks to succeed; there, however, measurements suggest that \agns{}
capture the inconspicuousness and realizability objectives
more effectively. 
}

Using a detector did not thwart the \agn{} attacks: success
rates for dodging
and impersonation were similar to when a detector was not
used. However, using a detector reduced the
inconspicuousness of attacks (see~\secref{sec:userstudy}).


We further tested whether attackers can be more successful by using
eyeglasses of different shapes: we trained \agn{}s to generate
eyeglasses of six new shapes and tested them 
against \vggL{} and \openfaceL{}. Three of the new
shapes achieved comparable performance to the original shape (shown in
\figref{fig:digital_dodge}), but the overall success rates did not improve. 
Nevertheless, it would be useful to explore whether using variety of eyeglass shapes 
can enhance the inconspicuousness of attacks in practice.

\subsection{Attacks in the Physical Domain}
\label{sec:real}

Attackers in the physical domain do not have complete control over the
\dnn{}'s input: Slight changes in the attacker's pose, expression, distance
from the camera, and illumination may dramatically change the
concrete values of pixels. Practical attacks need to be robust against
such changes. We took three additional measures to make the attacks
more robust.

First, to train adversarial generators that emit images of eyeglasses
that lead to more than one of the attacker's images to be misclassified,
we used multiple images of the attacker in training the
generator. Namely, we set $X$ in~\algref{alg:agn} to be a collection
of the attacker's images. As a result, the generators learned to
maximize \dnnloss{} for different images of the attacker.


Second, to make the attacks robust to changes in pose, we trained
the adversarial generator to minimize \dnnloss{} over multiple images
of the attacker wearing the eyeglasses.
To align
the eyeglasses to the attacker's face, we created and printed a 3d
model of eyeglasses with frames that have the same 
silhouette as the 2d eyeglasses emitted by the generator
(using code from GitHub~\cite{Jenny2DTo3D}).
We added tracking markers---specifically positioned green dots---to the
3d-printed eyeglasses. The attacker wore the eyeglasses when
capturing training data for the generator. We then used the markers to
find a projective alignment, $\theta_{x}$, of the eyeglasses
emitted by the generator to the attacker's pose in each
image. The generator was subsequently trained to minimize
$\Loss_{F}\big(x+\theta_{x}(G(z))\big)$ for different images of the
attacker ($x \in X$).

Third, to achieve robustness to varying illumination conditions, we
modeled how light intensity (luminance) affects eyeglasses and
incorporated the models in \agn{} training. Specifically, we
used the Polynomial Texture Maps approach~\cite{Malzbender01PTM} to
estimate degree-3  polynomials that map eyeglasses' RGB values
under baseline luminance to values under
a specific luminance. In the forward pass of \algref{alg:agn}, before
digitally attaching eyeglasses to an attacker's image of certain luminance,
we mapped the eyeglasses' colors to match the image's luminance.
In the backward pass, the errors were back-propagated through the polynomials
before being back-propagated through the generator to adjust its weights. In this way, the texture-map polynomials enabled
us to digitally estimate the effect of lighting on the eyeglasses.

\parheading{Experiment setup}
To evaluate the physically realized attacks, three subjects  from our
team acted as attackers: \subject{A} (the $1^\mathit{st}$ author),
a Middle-Eastern male in his mid-20s; \subject{B} (the $3^\mathit{rd}$
author), a white male in his 40s; and \subject{C} (the $2^\mathit{nd}$
author), a South-Asian female in her mid-20s. Each subject attempted
both dodging and impersonation against each of the four \dnn{}s
(which were trained to recognize them, among others).
The data used for training and evaluating the physically realized
attacks were collected from a room on Carnegie Mellon University's Pittsburgh
campus using a Canon T4i camera. To control the illumination more accurately, we
selected a room with a ceiling light but no windows on exterior walls. 

In a first set of experiments, we evaluated the attacks under varied
poses. To train the adversarial generators, 
we collected 45 images of each attacker (the set $X$ in \algref{alg:agn})
while he or she stood a fixed distance from the camera, kept a neutral
expression, and moved his or her head up-down, left-right, and in a circle.
Each generator was trained for at most one epoch, and training stopped
earlier if the generator could emit
eyeglasses that, for dodging, led the mean probability of
the correct class to fall below 0.005, or, for impersonation,
led the mean probability of the target class to exceed 0.990.
For impersonation, we picked the target at random per attack. 

To physically realize the attacks, we printed selected eyeglass patterns
created by
the generator on Epson Ultra Premium Glossy paper, using a commodity
Epson XP-830 printer,
and affixed them to the 3d-printed eyeglasses. Since each generator can emit
a diverse set of eyeglasses, we (digitally) sampled 48 outputs (qualitatively,
this amount seemed to capture the majority of patterns that the generators
could emit) and kept the most successful one for dodging or
impersonation in the digital environment (i.e., the one that led to the lowest
mean probability assigned the attacker or the highest mean probability assigned
to the target, respectively).

We evaluated the attacks by collecting videos of the attackers wearing the 3d-printed
eyeglasses with the adversarial patterns affixed to their front. Again, the attackers
were asked to stand a fixed distance from the camera, keep a neutral expression,
and move their heads up-down, left-right, and in a circle. 
We extracted every third frame from each video. This resulted in 75 frames, on
average, per attack. We then classified the extracted images using
the \dnn{}s targeted by the attacks.
For dodging, we measured success by the fraction
of frames that were classified as anybody but the attacker, and for impersonation
by the fraction of frames that were classified as the target.
In some
cases, impersonation failed---mainly due to the generated eyeglasses not being
realizable, as many of the pixels had extreme values (close to RGB=[0,0,0]
or RGB=[1,1,1]). In such cases, we attempted to impersonate another
(randomly picked) target.

We measured the head poses (i.e., pitch, yaw, and roll angles) of the
attackers in training images using a state-\-of-the-art tool~\cite{Baltruvsaitis16Openface2}.
On average, head poses covered 13.01$^{\circ}$ of pitch (up-down
direction), 17.11$^{\circ}$ of yaw (left-right direction), and
4.42$^{\circ}$ of roll
(diagonal direction). This is similar to the mean difference in head pose between
pairs of images randomly picked from the PubFig dataset (11.64$^{\circ}$ of pitch,
15.01$^{\circ}$ of yaw, and 6.51$^{\circ}$ of roll).

As a baseline to compare to, we repeated the dodging and impersonation attempts
using our prior attack~\cite{Sharif16AdvML}, referred to by \ccsattack{}.
\hlnew{Unlike experiments in the digital domain, we did not evaluate variants
  of the \ccsattack{} attack where additional clipping is performed, as our
experience in the digital domain showed that clipping harms the success
rate of attacks and fails to improve their inconspicuousness (see 
\secref{sec:userstudy}).}

In a second set of experiments, we evaluated the effects of changes to
luminance. To this end, we placed a lamp (with a 150$\mathit{W}$
incandescent light bulb) about 45$^{\circ}$ to the left of the attacker, and
used a dimmer to vary the overall illuminance between $\sim$110$\mathit{lx}$
and $\sim$850$\mathit{lx}$ (comparable to difference
between a dim corridor and a bright chain store interior~\cite{Luminance}). We crafted the
attacks by training the generator on 20 images of the attacker collected over
five equally spaced luminance levels. In training the generator, we used the
polynomial texture models as discussed above. For impersonation, we
used the same targets as in the first set of experiments.
We implemented the
eyeglasses following the same procedure as before, then collected 40
video frames
per attack, split evenly among the five luminance levels. In these experiments, the
attackers again stood a fixed distance from the
camera, but did not vary their pose.
For this set of experiments, we do not compare with our previous algorithm, as it
was not designed to achieve robustness to changing luminance, and
informal experiments showed that it performed poorly when varying the
luminance levels.

\begin{figure}[t]
\centering
    \begin{subfigure}[t]{0.25\textwidth}
    \centering
    \includegraphics[width=\textwidth,height=\textwidth]{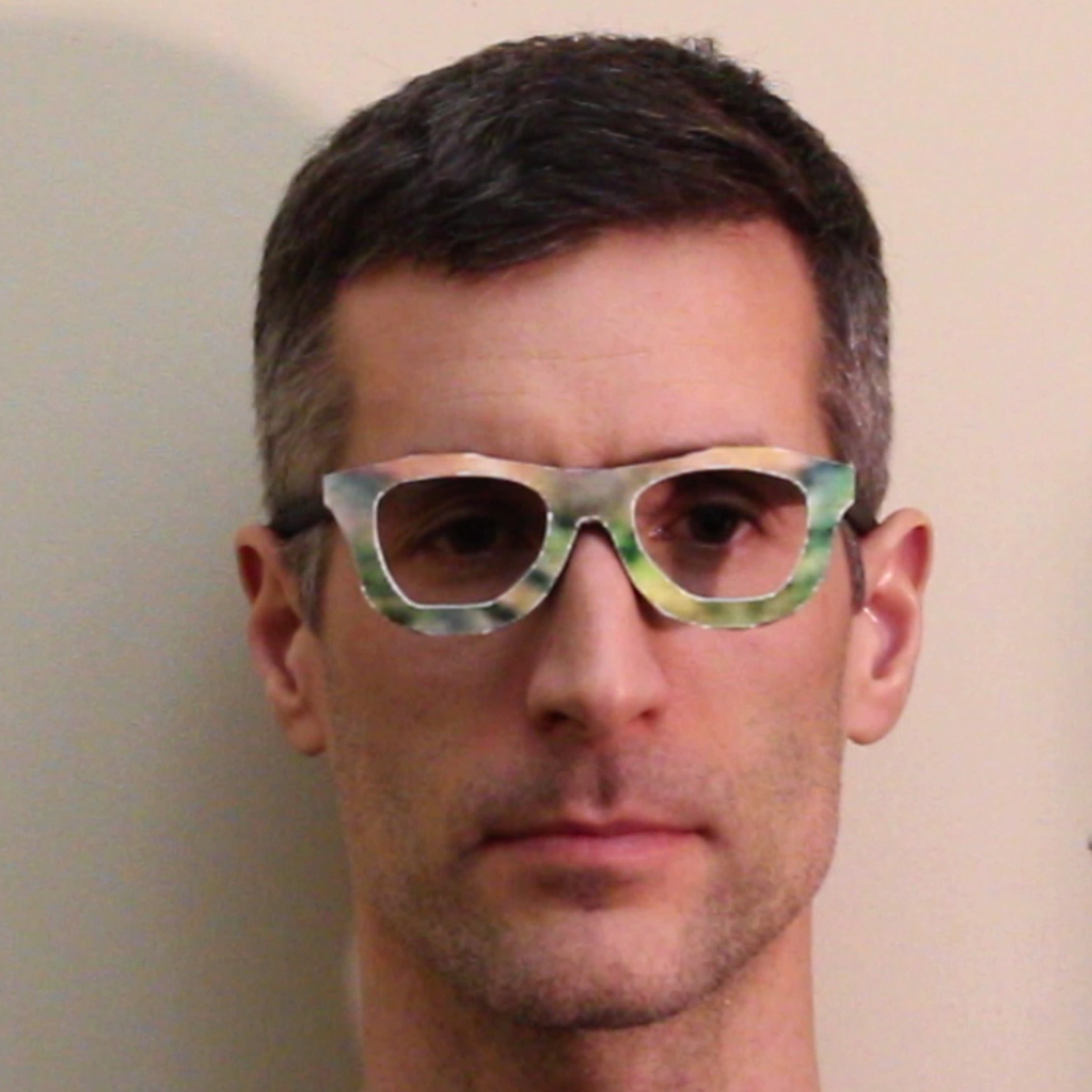}\\[0.2cm]
    \includegraphics[width=\textwidth,height=\textwidth]{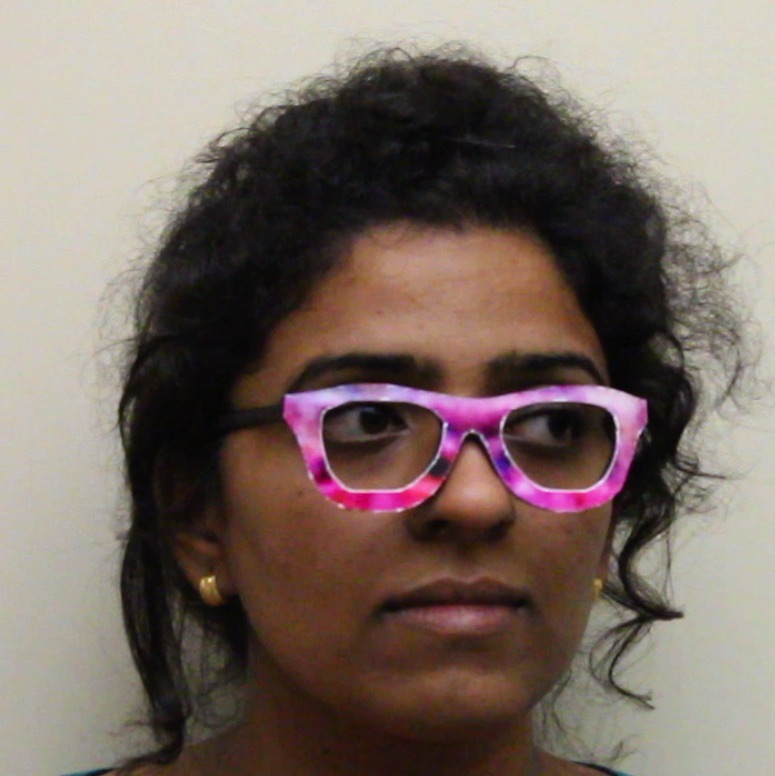}
    \caption{\label{fig:phys_dodge}}
  \end{subfigure}\hspace{0.2cm}
  \begin{subfigure}[t]{0.25\textwidth}
    \centering
    \includegraphics[width=\textwidth,height=\textwidth]{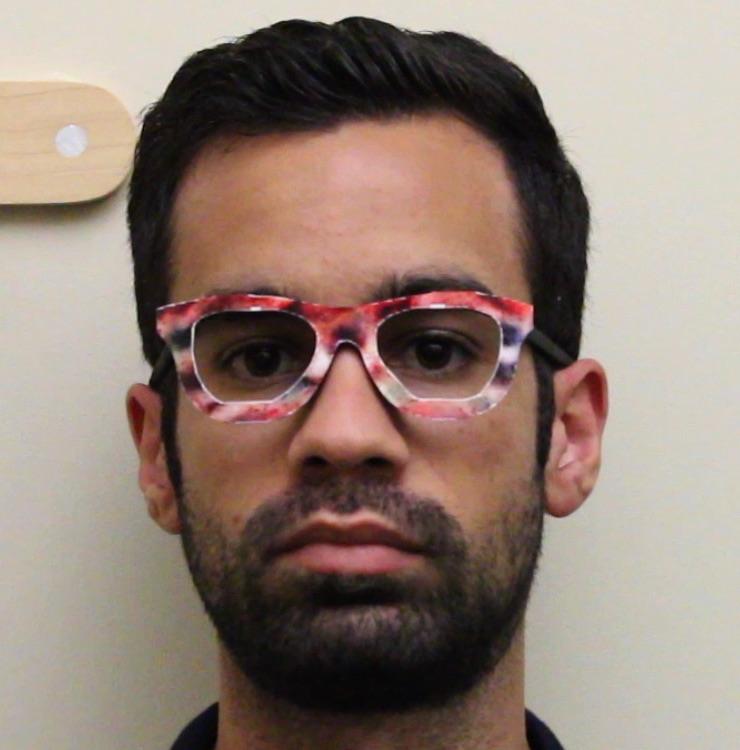}\\[0.2cm]
    \includegraphics[width=\textwidth,height=\textwidth]{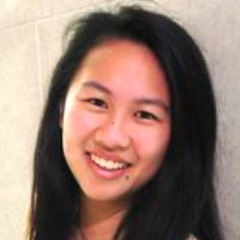}
    \caption{\label{fig:phys_imp1}}
  \end{subfigure}\hspace{0.2cm}
  \begin{subfigure}[t]{0.25\textwidth}
    \centering
    \includegraphics[width=\textwidth,height=\textwidth]{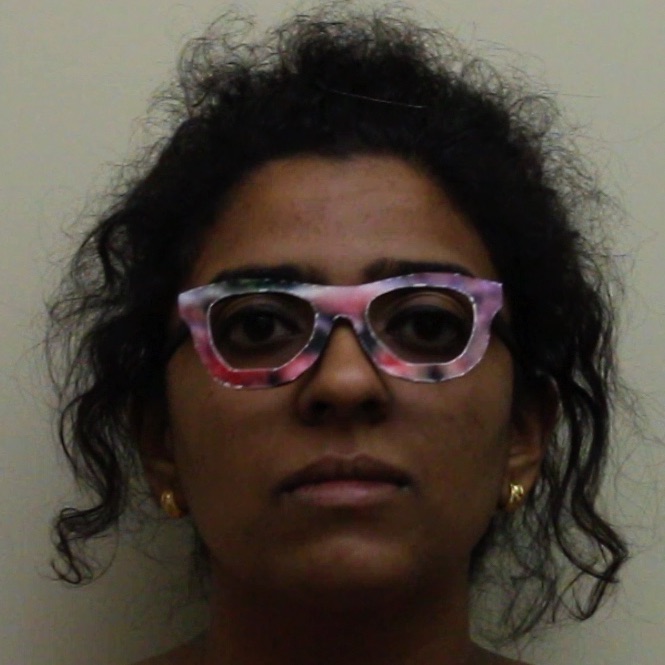}\\[0.2cm]
    \includegraphics[width=\textwidth,height=\textwidth]{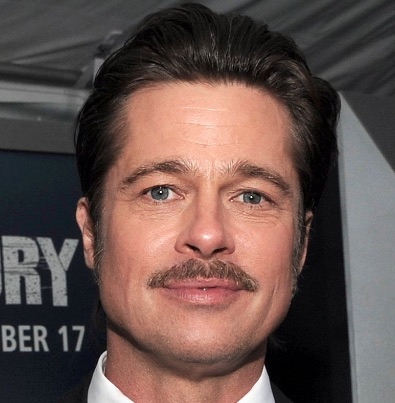}
    \caption{\label{fig:phys_imp2}}
  \end{subfigure}
\caption{\label{fig:physical_attacks} Examples of physically realized attacks.
	(a) \subject{B} (top) and \subject{C} (bottom) dodging against \openfaceL{}.
	(b) \subject{A} impersonating \subject{D} against \vggS{}.
	(c) \subject{C} impersonating actor Brad Pitt (by Marvin Lynchard /
	CC BY 2.0 / cropped from \url{https://goo.gl/Qnhe2X}) against \vggS{}.}
\end{figure}

\parheading{Experiment results} \tabref{tab:physical} summarizes our
results and \figref{fig:physical_attacks} shows examples of attacks in
the physical environment.

\begin{table*}[tb]
\begin{center}
\footnotesize
\tabcolsep=0.11cm
	\begin{tabular} { c | c | r r r r | c  c r r r r r }
	\toprule
	{} & {} & \multicolumn{4}{c|}{Dodging results} 
		& \multicolumn{7}{c}{Impersonation results} \\ 
	\emph{DNN} & \emph{Subject} 
		& \emph{\agn{}s} & $p(\text{sub.})$ & \emph{\ccsattack{}} & \emph{\agn{}s-L}
		& \emph{Target}  & \emph{Attempts} & \emph{\agn{}s} & \emph{HC} 
		& $p(\text{tar.})$ & \emph{\ccsattack{}} & \emph{\agn{}s-L}  \\ \midrule
	\vggS & \subject{A} & 100\% & 0.06 & 0\% & 93\%
			& \subject{D} & 1 & 100\%  & 74\% & 0.93 & 0\% & 80\% \\
	{} & \subject{B} & 97\% & 0.09 & 98\% & 100\%
			& Milla Jovovich & 2 & 88\%  & 0\% & 0.70 & 63\% & 100\% \\
	{} & \subject{C} & 96\% & 0.10 & 100\% & 100\%
			& Brad Pitt & 1 & 100\%  & 96\% & 0.98 & 100\% & 100\% \\ [3pt]
	\vggL & \subject{A} & 98\% & 0.17 & 0\% & 100\%
			 & Alicia Keys & 2 & 89\%  & 41\% & 0.73 & 0\% & 70\% \\
	{} & \subject{B} & 100\% & $<$0.01 & 87\% & 100\%
			 & Ashton Kutcher & 2 & 28\%  & 0\% & 0.22 & 0\% & 0\% \\ 
	{} & \subject{C} & 100\% & 0.03 & 82\% & 100\%
			& Daniel Radcliffe & 2 & 3\%  & 0\% & 0.04 & 0\% & 16\% \\ [3pt]
	\openfaceS & \subject{A} & 81\% & 0.40 & 0\% & 83\%
			& Brad Pitt & 2 & 28\%  & 23\% & 0.25 & 0\% & 50\% \\
	{} & \subject{B} & 100\% & 0.01 & 100\% & 100\%
			& Brad Pitt & 2 & 65\%  & 55\% & 0.58 & 43\% & 100\% \\ 
	{} & \subject{C} & 100\% & 0.01 & 100\% & 100\%
			 & \subject{D} & 1 & 98\%  & 95\% & 0.83 & 67\% & 38\% \\ [3pt]
	\openfaceL & \subject{A} & 100\% & 0.09 & 36\% & 75\%
			& Carson Daly & 1 & 60\%  & 0\% & 0.41 & 0\% & 73\% \\
	{} & \subject{B} & 97\% & 0.05 & 97\% & 100\%
			& Aaron Eckhart & 4 & 99\%  & 25\% & 0.83 & 92\% & 100\% \\
	{} & \subject{C} & 100\% & $<$0.01 & 99\% & 100\%
			& Eva Mendes & 2 & 53\%  & 39\% & 0.67 & 3\% & 9\% \\ [3pt]
	\bottomrule
	\end{tabular}
\caption{Summary of physical realizability experiments. 
	For dodging, we report the success rate of \agn{}s (percentage of misclassified video frames),
	the mean probability assigned to the correct class (lower is better), the
	success rate of the \ccsattack{} attack~\cite{Sharif16AdvML}, and the success rate
        of \agn{}s under 
	luminance levels higher than the baseline luminance level (\agn{}s-L). For impersonation, 
	we report the target (\subject{D} is 
	a member of our group, an Asian female in the early 20s),
	the number of targets attempted until succeeding, the success rate of \agn{}s
	(percentage of video frames classified as the target), the fraction of
	frames classified as the target with high confidence (HC; above the threshold which strikes a good balance
	between the true and the false positive rate), the mean probability assigned to the
	target (higher is better),
	the success rate of the \ccsattack{} attack~\cite{Sharif16AdvML}, and the success 
	rate under varied luminance levels excluding the baseline level (\agns{}-L).
	Non-adversarial images of the attackers were assigned to the correct class.
}
\label{tab:physical}
\end{center}
\end{table*}

In the first set of experiments we varied 
the attackers' pose. Most dodging attempts succeeded with all video
frames misclassified. Even in the worst attempt, 81\% of video frames
were misclassified. Overall, the mean probability assigned to the
correct class was at most 0.40, much below the thresholds discussed
in~\secref{sec:dnns}. For impersonation,
one to four
subjects had to be targeted before impersonations succeeded, with
an average of 
68\% of video frames (mis)classified as the targets
in successful impersonations. In two thirds of
these attempts, $>$20\% of frames were misclassified with high
confidence (again, using the thresholds from~\secref{sec:dnns}).
This suggests that even a conservatively
tuned system would likely be fooled by some attacks.

We found that physical-domain evasion attempts using \agn{}s were
significantly more successful than attempts using the \ccsattack{}
algorithm. The mean success rate of dodging
attempts was 45\% higher when using \agn{}s compared to prior work
(97\% vs.\ 67\%; a paired t-test shows that the difference
is statistically significant with $p=0.03$).
The difference in success rates
for impersonation was even larger. The mean success rate of impersonation
attempts was 126\% higher using \agn{}s compared to prior work (70\%
vs.\ 31\%; paired t-test shows that the difference is statistically
significant with $p<0.01$). Given these results, we believe that \agn{}s
provide a better approach to test the robustness of \dnn{}s 
against physical-domain attacks than the \ccsattack{} algorithm.


The second set of experiments shows that \agn{}s can generate attacks that
are robust to changes in luminance. On average, the dodging success rate
was 96\% (with most attempts achieving 100\% success rate),
and the impersonation success rate was 61\%. Both are comparable to
success rates under changing pose and fixed luminance. To evaluate
the importance of modeling luminance to achieve robustness,
we measured the success rate of \subject{A} dodging
against the four \dnn{}s \emph{without} modeling luminance
effects. This caused the
average success rate of attacks to drop from 88\% (the average success rate
of \subject{A} at dodging when modeling luminance) to 40\%
(marginally significant according to a t-test, with $p=0.06$). This suggests that
modeling the effect of luminance when training \agn{}s is essential to achieve
robustness to luminance changes.

\begin{table}[t]
\begin{center}
\small
  \begin{tabular} {l r r r}
	\toprule
	Factor & $\log(\mathit{odds})$ & $\mathit{odds}$ & \emph{p-value} \\ \midrule
	(intercept) 					& $<$0.01 & 1.00 & 0.96 \\[3pt]
	\textbf{is.143.subjects.dnn} 		& -0.48 & 0.62 & $<$0.01 \\
	\textbf{is.openface.dnn} 			& 6.34 & 568.78 & $<$0.01 \\
	\textbf{abs(pitch)}				& -0.06 & 0.94 & $<$0.01 \\
	\textbf{abs(yaw)}				& -0.06 & 0.94 & $<$0.01 \\
	abs(roll)						& 0.01 & 1.01 & 0.80 \\
	luminance					& 0.04 & 1.04 & 0.12 \\
	\textbf{non-printability}			& -1.09 & 0.34 & $<$0.01 \\[3pt]
	is.openface.dnn:luminance		& 0.38 & 1.48 & 0.14 \\
	\textbf{is.openface.dnn:abs(pitch)}  & 0.18 & 1.19 & $<$0.01 \\
	is.openface.dnn:abs(yaw)		& -0.08 & 0.92 & 0.06 \\
	\textbf{is.openface.dnn:abs(roll)}     & -0.62 & 0.54 & $<$0.01 \\
	\bottomrule
	\end{tabular}

  \caption{\label{tab:regression}
	Parameter estimates for the logistic regression model.
	Statistically significant factors are in boldface.
	}
\end{center}
\end{table}

Last, we built a mixed-effects logistic regression model~\cite{nlme2015} to analyze
how different factors, and especially head pose and luminance, affect the success
of physical-domain attacks. In the model, the dependent variable was whether
an image was misclassified, and the independent variables accounted for the absolute
value of pitch (up-down), yaw (left-right), and roll (tilt) angles of the head in the image
(measured with Baltru{\v{s}}aitis et al.'s
tool~\cite{Baltruvsaitis16Openface2}); the luminance level (normalized to a
[0,4] range);
how close are the colors of the eyeglasses printed for the attack to colors that can be
produced by our printer (measured via the \textit{non-printability score} defined
in our prior work~\cite{Sharif16AdvML}, and normalized to a [0,1] range); the
architecture of the \dnn{} attacked (VGG or OpenFace); and the size of the
\dnn{} (10 or 143 subjects). The model also accounted for the interaction
between angles and architecture, as well as the luminance and architecture.

To train the model, we used all the images we collected to test the attack in
the physical domain. The model's $R^2$ is 0.70 (i.e., it explains 70\% of the
variance in the data), indicating a good fit. The parameter estimates are shown
in \tabref{tab:regression}. Luminance is not a statistically significant
factor---i.e., the \dnn{}s were equally likely to misclassify the images under the
different luminance levels we considered. In contrast, the face's pose has a significant
effect on misclassification. For the VGG networks, each degree of pitch or
yaw away from 0$^{\circ}$ reduced the likelihood of success by 0.94,
on average. Thus, an attacker who faced the camera at a pitch or yaw of
$\pm$10$^{\circ}$ was about 0.53 times less likely to succeed than when
directly facing the camera. Differently from the VGG networks, for the
OpenFace networks each degree
of pitch away from 0$^{\circ}$ increased the likelihood of success by
1.12, on average. Thus, an attacker facing
the camera at a pitch of $\pm$10$^{\circ}$ was about 3.10 times more likely to
succeed than when directly facing the camera. Overall, these results
highlight the attacks' robustness to changes in luminance, as well as
to small changes in pose away from frontal.

\subsection{Universal Dodging Attacks}
\label{sec:universal}

We next show that a small number of adversarial
eyeglasses can allow successful dodging for the majority of
subjects, even when images of those subjects are not used in training
the adversarial generator.


\begin{algorithm}
 \small
 \caption{Universal attacks (given many subjects)}
 \label{alg:universal}

  \SetKwData{Left}{left}\SetKwData{This}{this}\SetKwData{Up}{up}
  \SetKwFunction{Union}{Union}\SetKwFunction{FindCompress}{FindCompress}
  \SetKwInOut{Input}{Input}\SetKwInOut{Output}{Output}

  \Input{$X$, \gen{}, \discrim{}, \dnnFunc{}, $\mathit{dataset}$, $Z$, $N_e$, $s_b$, 
	$\kappa$, $s_{c}$}
  \Output{ $\mathit{Gens}$ \tcp*[h]{a set of generators} }
  \BlankLine

  $\mathit{Gens}$ $\leftarrow \{\}$\;
  $\mathit{clusters}$ $\leftarrow$ clusters of size $s_{c}$ via \textit{k-means++}\;
  \For{$\mathit{cluster}\in \mathit{clusters}$}{
    $G \leftarrow \mathit{Alg1}(\mathit{cluster}, G, D, F, \mathit{dataset}, Z, N_e, s_b, \kappa)$\;
    $\mathit{Gens} \leftarrow \mathit{Gens} \cup \{G\}$\;
  }
  \Return{$\mathit{Gens}$}\;
\end{algorithm}

We created the universal attacks by training the generator in \algref{alg:agn} on a
set of images of different people. Consequently, the generator learned to
emit eyeglasses that caused multiple people's images to be misclassified, not only
one person's. We found that when the number of subjects was large, the generator
started emitting conspicuous patterns that did not resemble real eyeglasses.
For such cases, we used \algref{alg:universal}, which builds on \algref{alg:agn}
to train several adversarial generators, one per cluster of similar subjects.
\algref{alg:universal} uses \emph{k-means++}~\cite{Arthur07KMeans} to create
clusters of size $s_c$. Clustering was performed in Euclidian space using the features extracted
from the base \dnn{}s (4096-dimensional features for VGG, and 128-dimensional
features for OpenFace; see \secref{sec:dnns}).
The result was a set of generators that create
eyeglasses that, cumulatively, (1) led to the misclassification of a large fraction of
subjects and (2) appeared more inconspicuous (as judged by members of our team)
than when training on all subjects combined. The key insight behind the algorithm is that
it may be easier to find inconspicuous universal adversarial eyeglasses for similar subjects
than for vastly different subjects.


\parheading{Experiment setup} We tested the universal attacks against
\vggL{} and \openfaceL{} only, as the other \dnn{}s were trained with
too few subjects to make meaningful conclusions.
To train and evaluate the generators, we selected two images
for each of the subjects the \dnn{}s were trained on---one image
for training and one image for testing. To make dodging more
challenging, we selected the two images that were classified correctly
with the highest confidence by the two networks. Specifically, we
selected images such that the product of the probabilities both
\dnn{}s assigned to the correct class was the highest among all the
available images.

To
explore how the number of subjects used to create the universal
attacks affected performance, we varied the number of
(randomly picked) subjects with whose images we trained the
adversarial generators.
We averaged the success rate
after repeating the process five times (each time selecting a random
set of subjects for training). When using $\ge 50$ subjects for the
universal attacks, we used \algref{alg:universal} and set the cluster
size to 10.

Additionally, we explored how the number of adversarial eyeglasses affected
the success of the attack. We did so by generating
100 eyeglasses from each trained generator or set of generators 
and identifying the subsets (of varying size) that led the largest fraction of
images in the test set to be misclassified.
Finding the optimal subsets is NP-hard, and so we used an
algorithm that provides a $(1-{1\over e})$-approximation of the optimal
success rate~\cite{Nemhauser78Submodular}.
\nospace{This is essentially a
\emph{maximum coverage problem}, which is
NP-hard~\cite{Hochbaum96NPApprox}. Therefore, we used a procedure
based on submodular function maximization to select the set of
eyeglasses~\cite{Nemhauser78Submodular}.  The algorithm starts from
selecting the eyeglasses that maximize the success rate (i.e., they
led to the misclassification of the largest fraction of images from
the test set).  It then proceeds iteratively, increasing the
set of eyeglasses by picking the pair that leads to the
misclassification of the largest fraction of test images that have
not yet been misclassified.
This algorithm provides a $(1-{1\over e})$-approximation of the optimal
success rate~\cite{Nemhauser78Submodular}.}

\begin{figure}[tbh]
	\centering
	\begin{subfigure}[b]{0.49\textwidth}
		\centering
		\includegraphics[width=0.98\columnwidth]{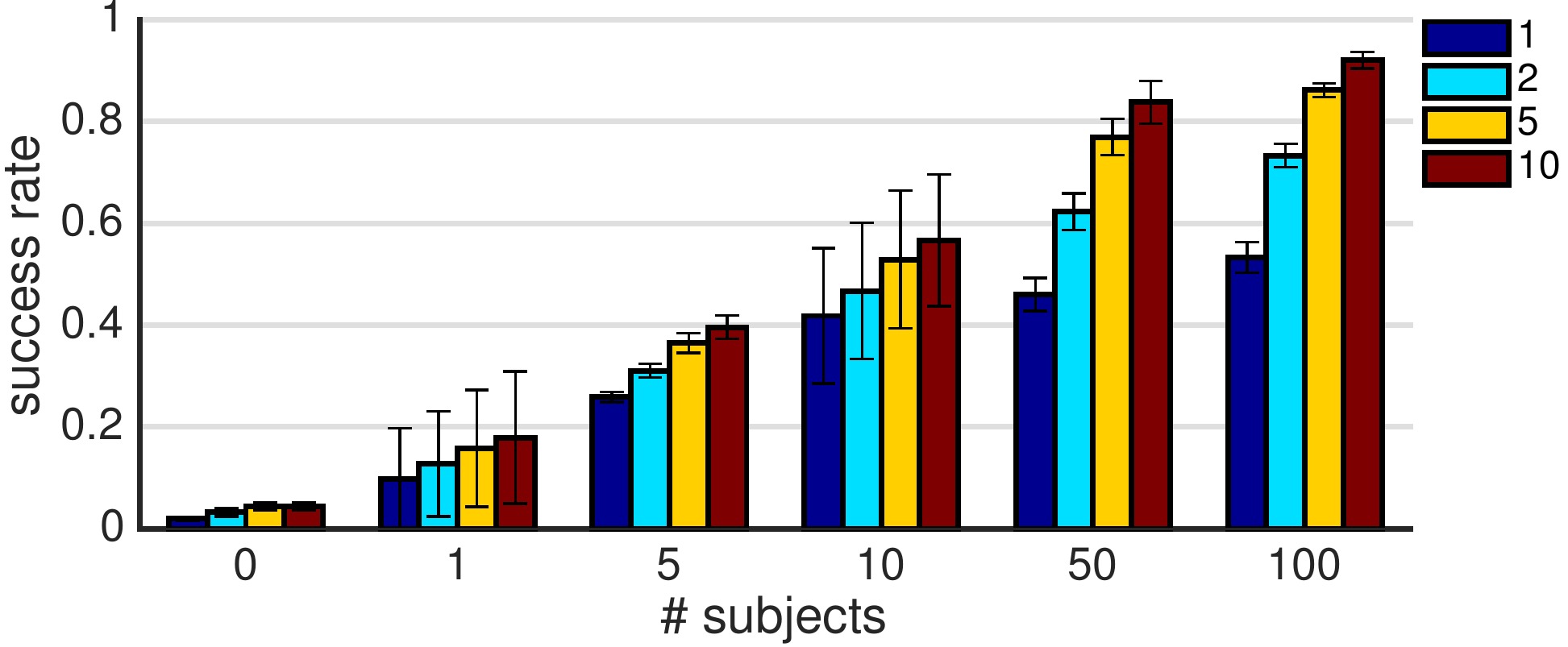}
	\caption{Universal dodging against \vggL{}.}\label{fig:vgg_univ}
	\end{subfigure}
	\begin{subfigure}[b]{0.49\textwidth}
		\centering
		\includegraphics[width=0.98\columnwidth]{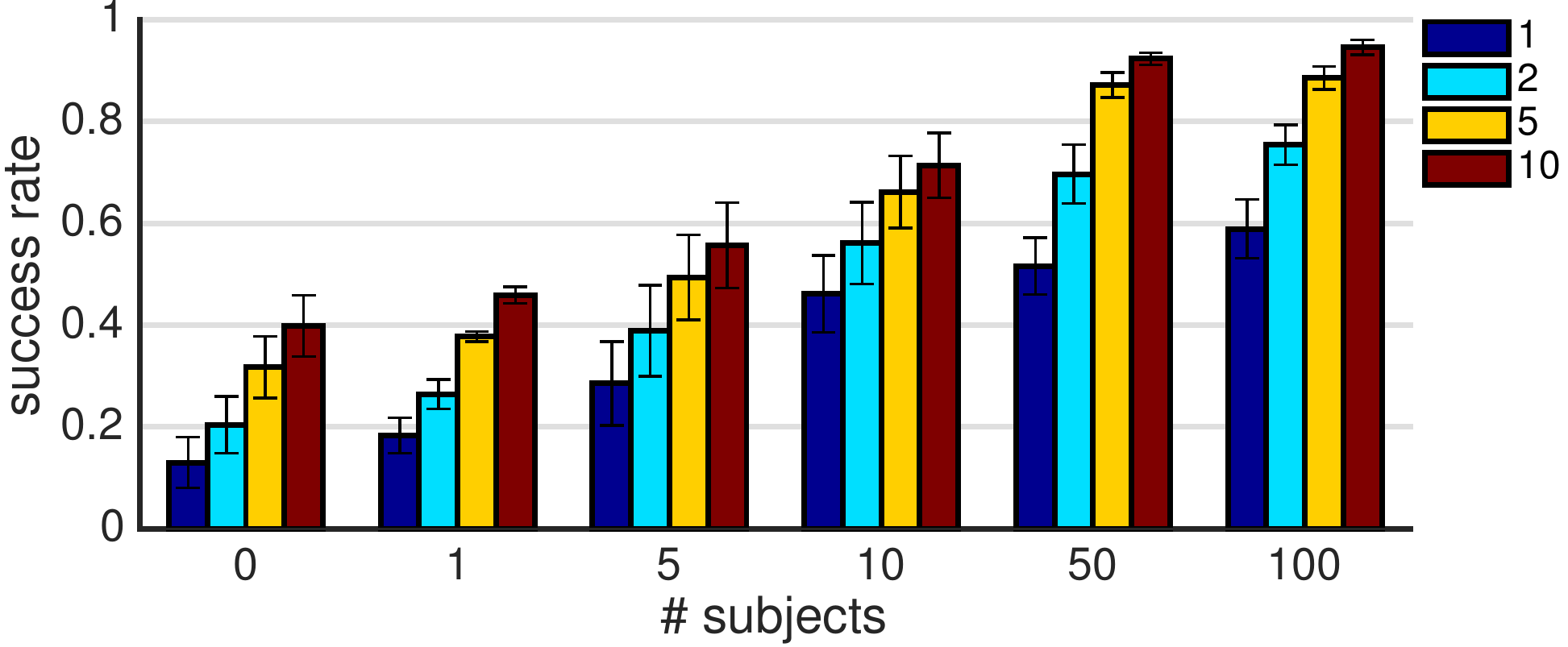}
                \caption{Universal dodging against \openfaceL{}.}
	\label{fig:openface_univ}
	\end{subfigure}
	\caption{Universal dodging against \vggL{} and \openfaceL{}. The x-axis shows
		the number of subjects used to train the adversarial generators. When the number of
		subjects is zero, a non-adversarial generator was used. The y-axis shows the mean fraction
		of images misclassified (i.e., the dodging success rate). The whiskers on the
		bars show the standard deviation of the success rate, computed by repeating each
		experiment five times, each time with a different set of randomly picked subjects. The
		color of the bars denotes the number of eyeglasses used, as shown in the legend.
		We evaluated each attack
		using one, two, five, or ten eyeglasses. For example,
                the rightmost bar in \figref{fig:openface_univ}
                indicates that an \agn{} trained with images of 100
                subjects will generate eyeglasses such that 10 pairs
                of eyeglasses will allow approximately 94\% of
                subjects to evade recognition.
		For $\le 10$ subjects, \algref{alg:agn} was used
		to create the attacks.
                For 50 and 100 subjects, \algref{alg:universal}
		was used.}
	\label{fig:universal}
\end{figure}

\parheading{Experiment results} \figref{fig:universal} summarizes the
results. Universal attacks are indeed possible: generators
trained to achieve dodging using a subset of subjects produced
eyeglasses that led to dodging when added to images of subjects not
used in training. The effectiveness of dodging depends chiefly on
the number of subjects used in training and, secondarily, the number
of eyeglasses generated. In particular, training a generator (set)
on 100 subjects and using it to create 10 eyeglasses was sufficient to
allow 92\% of remaining subjects to dodge against \vggL{} and 94\% of
remaining subjects to dodge against \openfaceL. Even training on five
subjects and generating five eyeglasses was sufficient to allow more
than 50\% of the remaining users to dodge against either network.
 \openfaceL{} was particularly more
susceptible to universal attacks than \vggL{} when a small number of
subjects was used for training, likely due to its overall lower accuracy.

\subsection{Transferability of Dodging Attacks}
\label{sec:transfer}


Transferability of attacks has been shown to be effective in fooling models to
which adversaries do not have access
(e.g.,~\cite{Papernot17Blackbox}).
 In our case, although this is not an explicit goal of our attacks, attackers with access to
one \dnn{} but not another may attempt to rely on transferability to dodge against
the second \dnn{}.
In this section, we explore whether dodging against \dnn{}s of one architecture
leads to successful dodging against \dnn{}s of a different architecture. 

\begin{table}
\centering
\small
    \begin{subtable}{.495\columnwidth}
      \centering
\begin{tabular}{ l | c c }  
	\backslashbox{From}{To} &	\vggS{} 	&	\openfaceS{} 	\\ \hline
	\vggS{}				& 	-  	 &	63.33\% \\
	\openfaceS{}			& 10.00\% &	-		\\
\end{tabular}
    \end{subtable}
    \begin{subtable}{.495\columnwidth}
\begin{tabular}{ l | c c } 
	\backslashbox{From}{To}  &	\vggL{} 	&	\openfaceL{} 	\\ \hline
	\vggL{}				& 	-		&	88.33\% 	\\
	\openfaceL{}			& 11.67\%	&	-	 	 \\
\end{tabular}
    \end{subtable}
    \caption{Transferability of dodging in the digital domain.
	Each table shows how likely it is for a generator used for dodging against
	one network (rows) to succeed against another network (columns).}
\label{tab:transfer_digital}
\end{table}

\begin{table}
\centering
\small
    \begin{subtable}{0.495\columnwidth}
      \centering
\begin{tabular}{ l | c c } 
	\backslashbox{From}{To} &	\vggS{} 	&	\openfaceS{} 	\\ \hline
	\vggS{}				& 	-	 	&	43.84\% 	\\
	\openfaceS{}			& 27.77\%	 	&	-	\\
\end{tabular}
    \end{subtable}
    \begin{subtable}{0.495\columnwidth}
\begin{tabular}{ l | c c } 
	\backslashbox{From}{To} &	\vggL{} 	&	\openfaceL{}  \\ \hline
	\vggL{}				& 	-	&	51.78\% 	  	\\
	\openfaceL{}			&  19.86\%		&	-			\\
\end{tabular}
    \end{subtable}
    \caption{Transferability of dodging in the physical domain. We classified
	the frames from the physically realized attacks using \dnn{}s different from the ones
	for which the attacks were crafted. Each table shows how likely it is for frames that
	successfully dodged against one network (rows) to succeed against another network
	(columns).
    }
\label{tab:transfer_physical}
\end{table}

Using the data from \secref{sec:digital}, we first tested whether
dodging in the digital environment successfully transferred between
architectures (see \tabref{tab:transfer_digital}). We found that
attacks against the OpenFace architecture successfully fooled the VGG
architecture in only a limited number of attempts (10--12\%). In
contrast, dodging against VGG led to successful dodging against
OpenFace in at least 63\% of attempts.

Universal attacks seemed to transfer between architectures with
similar success. Using attacks created with 100 subjects and 10
eyeglasses from \secref{sec:universal}, we found that 82\%
($\pm 3\%$ standard deviation) of attacks transferred from \vggL{}
to \openfaceL{}, and 26\% ($\pm 4\%$ standard deviation)
transferred in the other direction.

The transferability of dodging attacks in the physical environment between
architectures followed a similar trend (see \tabref{tab:transfer_physical}).
Successful attacks transferred less successfully from the OpenFace networks
to the VGG networks (20--28\%) than in the other direction (44--52\%).

\subsection{A Study to Measure Inconspicuousness}
\label{sec:userstudy}

\parheading{Methodology}
To evaluate inconspicuousness of eyeglasses generated by \agns{}
we carried out an online user
study. Participants were told that we were developing an  
algorithm for designing eyeglass patterns, shown a set of eyeglasses,
and asked to label each pair as either algorithmically generated or
real. Each participant saw 15 ``real'' and 15 attack
eyeglasses in random order. All eyeglasses were the same shape and
varied only in their coloring. The ``real'' eyeglasses were ones used for
pretraining the \agns{} (see \secref{sec:train-data}).
The attack eyeglasses were generated using either \agns{}, the \ccsattack{}
attack, \hlnew{or the \ccsclipped{} attack}.

Neither ``real'' nor attack eyeglasses shown to participants were
photo-realistically or three-dimensionally rendered. 
So, we consider attack glasses to have been
inconspicuous to participants not if they were uniformly rated as real
(which even ``real'' glasses were not, particularly when attack glasses
were inconspicuous),
but rather if the rate at which participants deemed them as real
does not differ significantly regardless of whether they are ``real''
eyeglasses or attack eyeglasses. 

Given
two sets of eyeglasses (e.g., a set of attack glasses and a set of
``real'' glasses), we tested whether one is more inconspicuous
via the $\chi^2$ test of independence~\cite{Mchugh13Chi2},
and conservatively corrected for multiple comparisons
using the Bonferroni correction.
We compared the magnitude of differences using the odds-ratio measure:
the odds of eyeglasses in the first group being marked as real divided by
the odds of eyeglasses in the second group being marked as real.
\hlnew{The higher (resp. lower) the odds ratios are from 1, the higher
  (resp. lower) was the likelihood that eyeglasses from the first group were
  selected as real compared to eyeglasses from the second group.}


We recruited 301 participants in the U.S.\ through the Prolific
crowdsourcing service.\footnote{\url{https://prolific.ac}}
Their ages ranged from 18 to 73, with a median of 29.
51\% of participants specified being female and 48\% male
(1\% chose other or did not answer). 
Our study took 3 minutes to complete on average and participants were
compensated \$1.50. The study design was approved by Carnegie Mellon University's
ethics review board.


\begin{table}[tb]
\begin{center}
\small
  
\tabcolsep=0.11cm
  \begin{tabular}{l @{\hspace{2mm}} l @{\hspace{2mm}} r @{\hspace{2mm}} r}
    \toprule
    && \emph{Odds}\\
    \multicolumn{2}{l}{\emph{Comparison (group 1 vs.\ group 2)}} & \emph{ratio} & \emph{p-value} \\ \midrule
    Real (61\%) & All \agns{} (47\%) & 1.71 & $<$0.01\\
    \agns{} digital (49\%) & \ccsattack{} digital~\cite{Sharif16AdvML} (31\%) & 2.19 & $<$0.01\\
    \agns{} digital (49\%) & \ccsclipped{} digital (30\%) & 2.24 & $<$0.01\\
    \agns{} physical (45\%) & \ccsattack{} physical~\cite{Sharif16AdvML} (34\%) & 1.59 & $<$0.01\\[3pt]
    \agns{}:\\
\    digital (49\%) & physical (45\%) & 1.19 & 0.26 \\
\    digital (49\%) & digital with detector (43\%) & 1.28 & 0.02 \\
\    digital dodging (52\%) & universal dodging (38\%) & 1.80 & $<$0.01 \\
\bottomrule
  \end{tabular}

  \caption{\hlnew{Relative realism of selected
    sets of eyeglasses. For each two sets compared, we report in parentheses the fraction of
	eyeglasses per set that were marked as real by study
        participants, the odds ratios between the groups, and the p-value of the $\chi^2$ test of independence.
	E.g., odds ratio of 1.71 means that
        eyeglasses are $\times$1.71 as likely to be selected as real if
        they are in the first set than if they are in the second.}}
  \label{tab:user_study}
  \end{center}
\end{table}



\begin{figure}
\includegraphics[width=0.7\columnwidth]{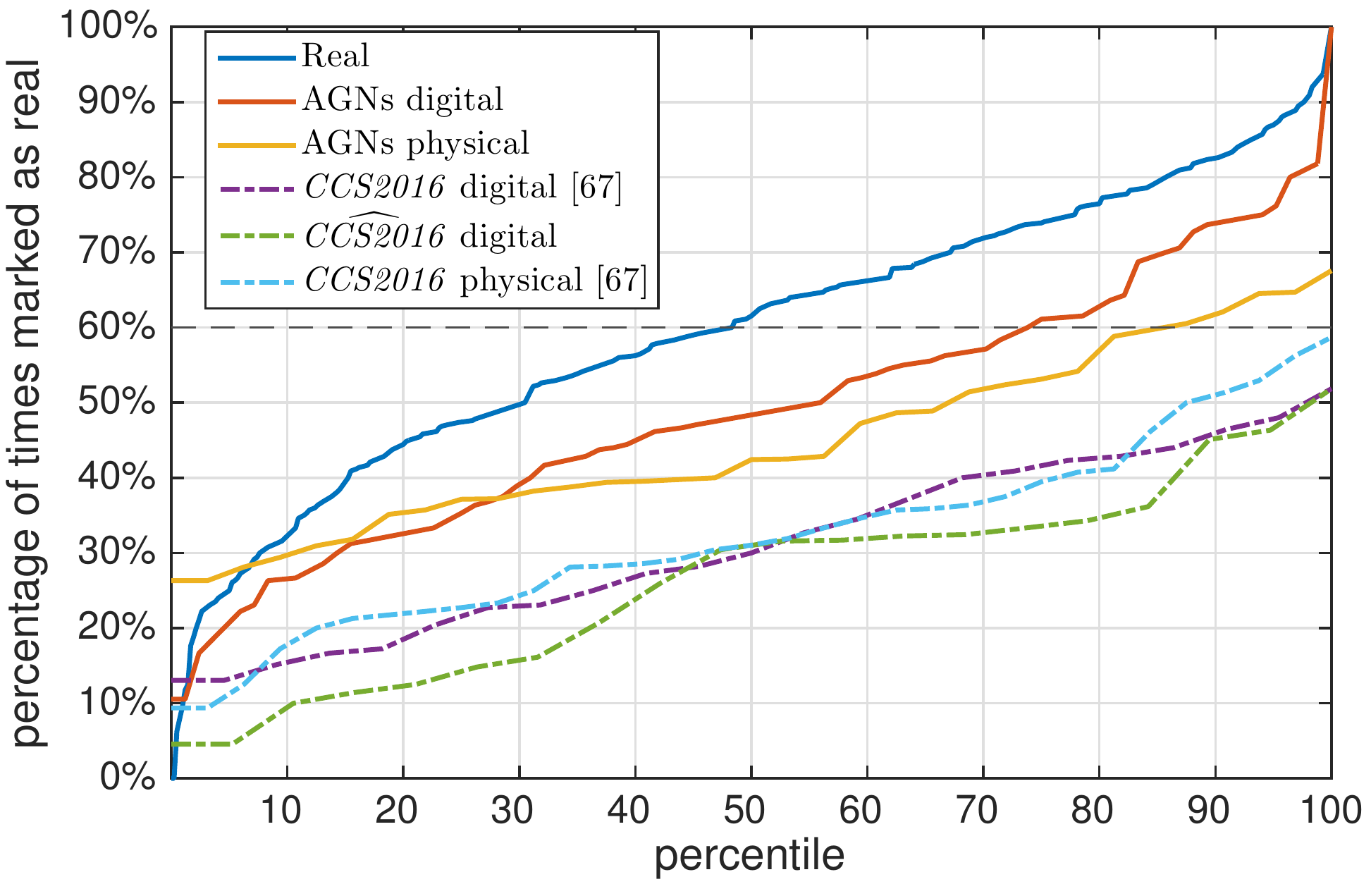}
\caption{\hlnew{The percentage of times in which eyeglasses from different
sets were marked as real. The horizontal 60\% line is highlighted to
mark that the top half of ``real'' eyeglasses were marked as real
at least 60\% of the time.}}
\label{fig:study}
\end{figure}

\parheading{Results}
\tabref{tab:user_study} and \figref{fig:study} show comparisons between various
groups of eyeglasses, as well as the percentage of time participants
marked different eyeglasses as real. ``Real'' eyeglasses were more realistic than
\agn{}-generated ones ($\times 1.71$ odds ratio). This is expected, given the
additional objectives that attack eyeglasses are required to achieve.
\hlnew{However, \agn{}s were superior to other attacks. Both for digital and
  physical attacks, eyeglasses created by \agn{}s were more realistic than those
  created by our previous \ccsattack{} attack~\cite{Sharif16AdvML}
  ($\times 2.19$ and $\times 1.59$ odds ratio, respectively). Even limiting the
  max-norm of the perturbations did not help---\agn{}s generated eyeglasses that
  were more likely to be selected as real than the \ccsclipped{} attack
  ($\times 2.24$ odds ratio).}

Perhaps most indicative of inconspicuousness in practice is that many
\agn{}-generated eyeglasses were as realistic as ``real'' eyeglasses.
The most inconspicuous 26\% of eyeglasses emitted by \agns{} for
digital-environment attacks were on average deemed as real as the
most inconspicuous 50\% of ``real'' eyeglasses; in each case
participants marked these eyeglasses as real $>$60\% of the time.
Physical attacks led to less inconspicuous eyeglasses; however, the
14\% most inconspicuous were still marked as real at least 60\% of the
time (i.e., as real as the top 50\% of ``real'' eyeglasses).

Other results match intuition---the more difficult the attack, the
bigger the impact on conspicuousness. Digital attack glasses
that do not try to fool a detector are less conspicuous than ones that
fool a detector ($\times 1.28$ odds ratio), and individual dodging is
less conspicuous than universal dodging ($\times 1.80$ odds ratio).
\hlnew{Digital attack glasses had higher odds of being selected as real than
  physical attack glasses ($\times 1.19$ odds ratio), but the differences
  were not statistically significant (\emph{p-value}=0.26).}

\subsection{\agn{}s Against Digit Recognition}
\label{sec:mnist}

We next show that \agns{} can be used in domains besides face
recognition. Specifically, we use \agns{} to fool
a state-of-the-art \dnn{} for recognizing digits, trained on the MNIST data\-set~\cite{MNIST},
which contains 70,000 28$\times$28-pixel images of digits. 

\parheading{Experiment setup} First, we trained a \dnn{} for digit
recognition using the architecture and training code of Carlini and
Wagner~\cite{Carlini17Robustness}. We trained the \dnn{} on 55,000
digits and used 5,000 for validation during training time.
The trained \dnn{} achieved 99.48\% accuracy on the test set of 10,000 digits.
Next, we pretrained 10 \gan{}s to generate digits, one for each digit.
Each generator was trained to map inputs randomly sampled
from $[-1,1]^{25}$ to 28x28-pixel images of digits. We again used the
Deep Convolutional \gan{} architecture~\cite{Radford15DCGAN}.
Starting from the pretrained \gan{}s, we trained \agn{}s using a variant of~\algref{alg:agn} to
produce generators that emit images of digits that
simultaneously fool the discriminator to be real and are misclassified
by the digit-recognition \dnn{}.

\hlnew{Unlike prior attacks, which typically attempted to minimally perturb
  specific benign inputs to cause misclassification 
  (e.g.,~\cite{Carlini17Robustness,Engstrom17AdvTrans,Gdfllw14ExpAdv,Papernot16Limitations,Poursaeed17GANAttack,Xiao18GANAttack}),
  the attack we propose does not assume that a benign input is
  provided, nor does it attempt to produce an attack image minimally
  different from a benign image. 
   Hence, a comparison with
   prior attacks would not be meaningful.}

\begin{figure}[tbh]
\centering
\includegraphics[width=42pt]{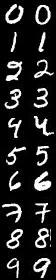}\hspace{12pt}
\includegraphics[width=42pt]{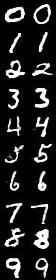}\hspace{12pt}
\includegraphics[width=42pt]{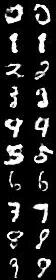}
\caption{An illustration of attacks generated via \agn{}s. Left: A random sample
	of digits from MNIST. Middle: Digits generated by the pretrained generator.
	Right: Digits generated via \agn{}s that are misclassified by the digit-recognition
	\dnn{}.}
\label{fig:mnist}
\end{figure}

\parheading{Experiment results} The \agns{} were able to output
arbitrarily many adversarial examples that appear comprehensible to
human observers, but are misclassified by the digit-recognition \dnn{} (examples
are shown in \figref{fig:mnist}). As a test, we generated 5,004 adversarial examples
that all get misclassified by the digit-recognition \dnn{}. The adversarial
examples were produced by first generating 600,000 images using the adversarial
generators (60,000 per generator). Out of all samples, the ones that were
misclassified by the \dnn{} (8.34\% of samples) were kept.
Out of these, only the digits that were
likely to be comprehensible by humans were kept: the automatic
filtering process to identify these involved
computing the product of the discriminator's output (i.e., how realistic the images
were deemed by the discriminator) and the probability assigned by the digit-recognition
\dnn{} to the correct class, and keeping the 10\% of digits with the highest product.

Differently from traditional attacks on digit recognition (e.g.,~\cite{Carlini17Robustness}),
these attack images are not explicitly designed for minimal deviation from specific
benign inputs; rather, their advantage is that they can be substantially different (e.g.,
in Euclidean distance) from training images. We measured the diversity of images
by computing the mean Euclidean distance between pairs of digits of the same
type; for attack images,
the mean distance was 8.34, while for the training set it was 9.25.

A potential way
\agns{} can be useful in this domain is adversarial training. For instance, by
augmenting the training set with the 5,004 samples, one can extend it
by almost 10\%. This approach can also be useful for visualizing inputs that
would be misclassified by a \dnn{}, but are otherwise not available in the training
or testing sets.

\section{Discussion and Conclusion}
\label{sec:conclusion}


In this paper we contributed a methodology that we call
\textit{adversarial generative nets} (\agns) to generate adversarial
examples to fool \dnn{}-based classifiers while meeting additional
objectives.
We focused on objectives imposed by the need to
physically realize artifacts that, when captured in an image,
result in misclassification of the image.  Using the physical
realization of eyeglass frames to fool face recognition as our driving
example, we demonstrated the use of \agn{}s to improve
robustness to changes in imaging conditions (lighting, angle, etc.)
and even to specific defenses; inconspicuousness to human onlookers;
and scalability in terms of the number of adversarial objects
(eyeglasses) needed to fool DNNs in different contexts.  \agn{}s generated
adversarial examples that improved upon prior work in all of these
dimensions, and did so using a general methodology.

Our work highlights a number of features of \agn{}s.  They are
flexible in their ability to accommodate a range of objectives,
including ones that elude precise specification, such as 
inconspicuousness. In principle, given an objective that can be
described through a set of examples, \agn{}s can be trained to emit
adversarial examples that satisfy this objective. Additionally, 
\agns{} are general in being applicable to various
domains, which we demonstrated by training \agn{}s to fool classifiers
for face and (handwritten) digit recognition. We expect that they
would generalize to other applications, as well. For example, one may
consider using \agn{}s to fool \dnn{}s for street-sign recognition by training
the generator to emit adversarial examples that resemble
street-sign images collected from the internet.

One advantage of \agn{}s over other attack methods
(e.g.,~\cite{Szegedy13NNsProps,Gdfllw14ExpAdv}) is that they can
generate multiple, diverse, adversarial examples for a given benign sample.
A diverse set of adversarial examples can be useful for evaluating
the robustness of models. Moreover, such adversarial examples may be
used to defend against attacks (e.g., by incorporating them in adversarial
training~\cite{Kurakin16AdvTrain}).

\section*{Acknowledgments}

The authors would like to thank Cara Bloom for her comments on an
early draft of this paper.  This work was supported in part by the
Multidisciplinary University Research Initiative (MURI) Cyber
Deception grant; by NSF grants 1801391 and 1801494; by the
National Security Agency under Award No.\ H9823018D0008;
by gifts from Google and Nvidia, and from Lockheed
Martin and NATO through Carnegie Mellon CyLab; and by a CyLab Presidential Fellowship and a
Symantec Research Labs Fellowship.

\bibliographystyle{plainnat}
\bibliography{cited}


\end{document}